\documentclass[letterpaper, 10 pt, conference]{ieeeconf}  % Comment this line out if you need a4paper

\usepackage{subfig}
\usepackage{times}
\usepackage{epsfig}
\usepackage{graphicx}
\usepackage{amsmath}
\usepackage{amssymb}
\usepackage{algorithm,algpseudocode,algorithmicx}
\usepackage{adjustbox}
\usepackage{multirow} 
\usepackage{etoolbox}
\usepackage{booktabs}
\usepackage{siunitx}

\usepackage{pdfpages}
\usepackage{pgfplots}
\pgfplotsset{compat=newest}
\usetikzlibrary{plotmarks}
\usetikzlibrary{arrows.meta}
\usepgfplotslibrary{patchplots}
\usepackage{grffile}
\usepackage{amsmath}

\PassOptionsToPackage{inline}{enumitem}

\usepackage[breaklinks=true,bookmarks=false]{hyperref}

\usepackage{tabularx,booktabs}
\usepackage{graphicx}
\IEEEoverridecommandlockouts                              % This command is only needed if 
\begin{document}

\title{Layouts from Panoramic Images with Geometry and Deep Learning}

\author{Clara Fernandez-Labrador, Alejandro Perez-Yus, Gonzalo Lopez-Nicolas, Jose J. Guerrero% <-this % stops a space
%\thanks{*This work was not supported by any organization}% <-this % stops a space
\thanks{Instituto de Investigaci\'on en Ingenier\'ia de Arag\'on (I3A), Universidad de Zaragoza, Spain. {\tt\small \{cfernandez, alperez, gonlopez, josechu.guerrero\} @unizar.es}}%
\thanks{This work was supported by Projects DPI2014-61792-EXP and DPI2015-65962-R (MINECO/FEDER, UE) and grant BES-2013-065834 (MINECO).}
}

\maketitle
\thispagestyle{empty}
\pagestyle{empty}

%%%%%%%%%%%%%%%%%%%%%%%%%%%%%%%%%%%%%%%%%%%%%%%%%%%%%%%%%%%%%%%%%%%%%%%%%%%%%%%%
\begin{abstract}

%Real world indoor scenarios can have different geometric shapes, contain many occlusions and be highly cluttered, that make them difficult to reproduce. Consequently, 
In this paper, we propose a novel procedure for 3D layout recovery of indoor scenes from single 360 degrees panoramic images.
With such images, all scene is seen at once, allowing to recover closed geometries. 
Our method combines strategically the accuracy provided by geometric reasoning (lines and vanishing points) with the higher level of data abstraction and pattern recognition achieved by deep learning techniques (edge and normal maps). Thus, we extract structural corners from which we generate layout hypotheses of the room assuming Manhattan world. The best layout model is selected, achieving good performance on both simple rooms (box-type) and complex shaped rooms (with more than four walls). Experiments of the proposed approach are conducted within two public datasets, SUN360 and Stanford (2D-3D-S) demonstrating the advantages of estimating layouts by combining geometry and deep learning and the effectiveness of our proposal with respect to the state of the art. 

% Real world indoor scenarios can have different geometric shapes, contain many occlusions and be highly cluttered. Consequently, this paper devises a novel flexible procedure for 3D layout recovery of indoor scenes from a single 360 degrees panoramic image achieving closed geometries without shape restrictions.
% This method combines strategically geometric reasoning on computer vision and deep learning techniques. 
% The precision provided by the geometry (lines and vanishing points extraction) together with the more human way of perception achieved by neural networks (edge and normal maps), allows us to carry out layout hypotheses based on structural corners and indoor scenes geometry obtaining a good performance on complex shaped rooms.
% Experimental tests of the proposed approach are conducted within two public datasets, SUN360 and Standford (2D-3D-S).
% We demonstrate the effectiveness of our proposal with respect to the state-of-the-art method, not only in terms of accuracy but also in handling more complex geometries, and also the advantages of estimating layouts by means of geometry and deep learning.  

\end{abstract}

%%%%%%%%%%%%%%%%%%%%%%%%%%%%%%%%%%%%%%%%%%%%%%%%%%%%%%%%%%%%%%%%%%%%%%%%%%%%%%%%

\section{INTRODUCTION}

\begin{figure}[t]
\captionsetup[subfloat]{farskip=2pt,captionskip=1pt}
\begin{center}
    \subfloat{\includegraphics[width=0.35\linewidth]{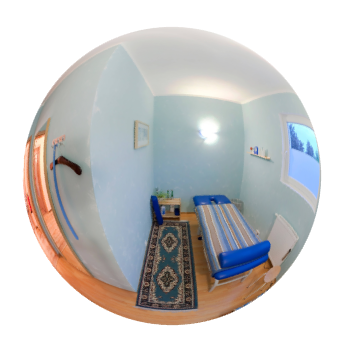}{\label{fig:i1}}} \hspace*{0.003\linewidth}
    \subfloat{\includegraphics[width=0.65\linewidth]{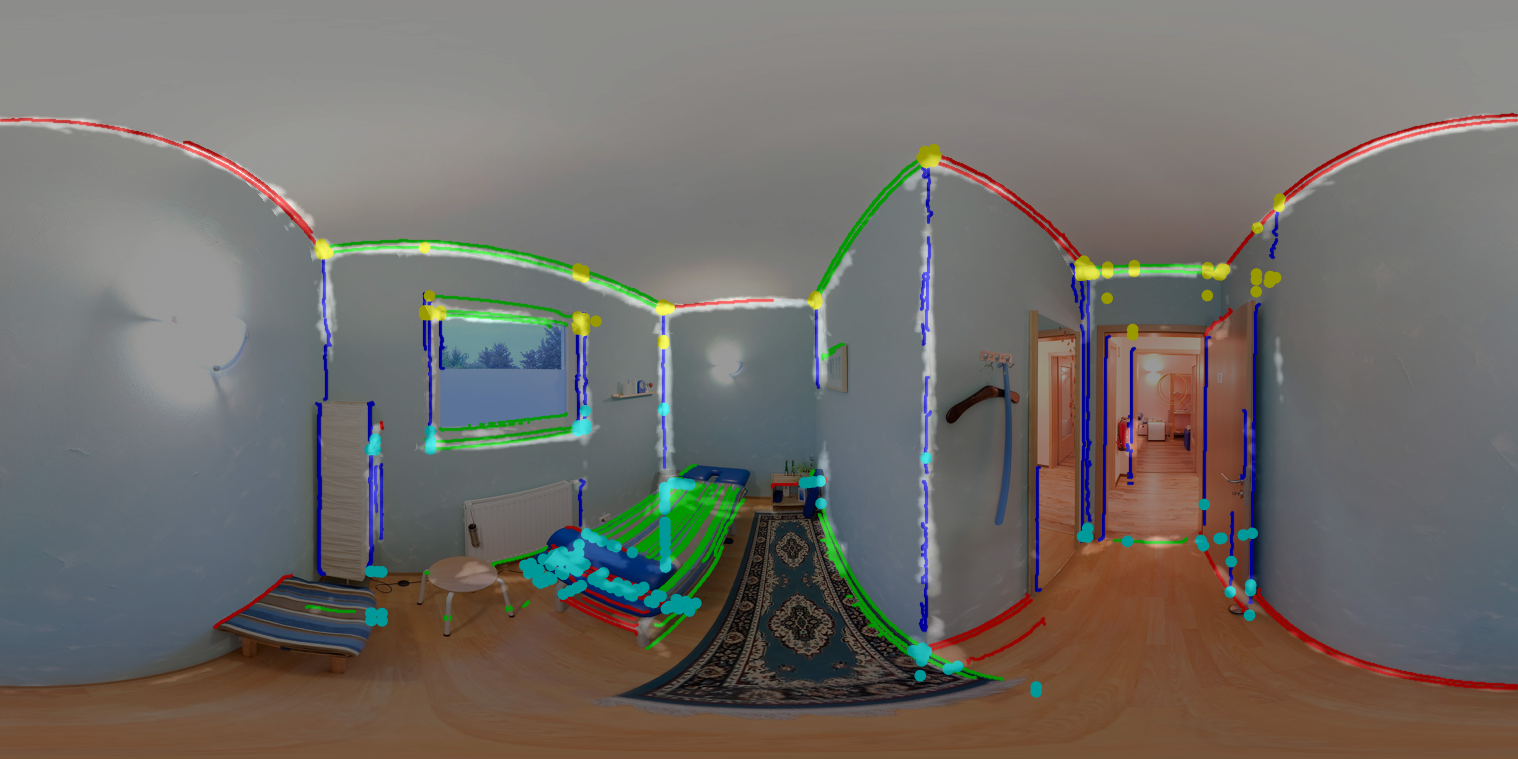}{\label{i2}}}\\
     \subfloat{\includegraphics[width=0.35\linewidth]{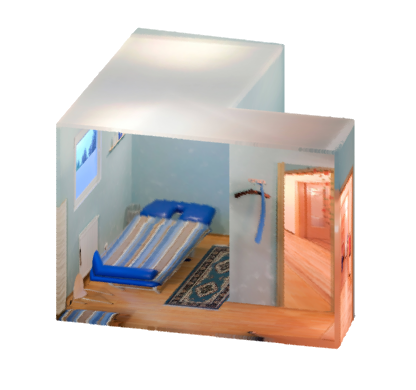}{\label{fig:i3}}} \hspace*{0.003\linewidth}
    \subfloat{\includegraphics[width=0.65\linewidth]{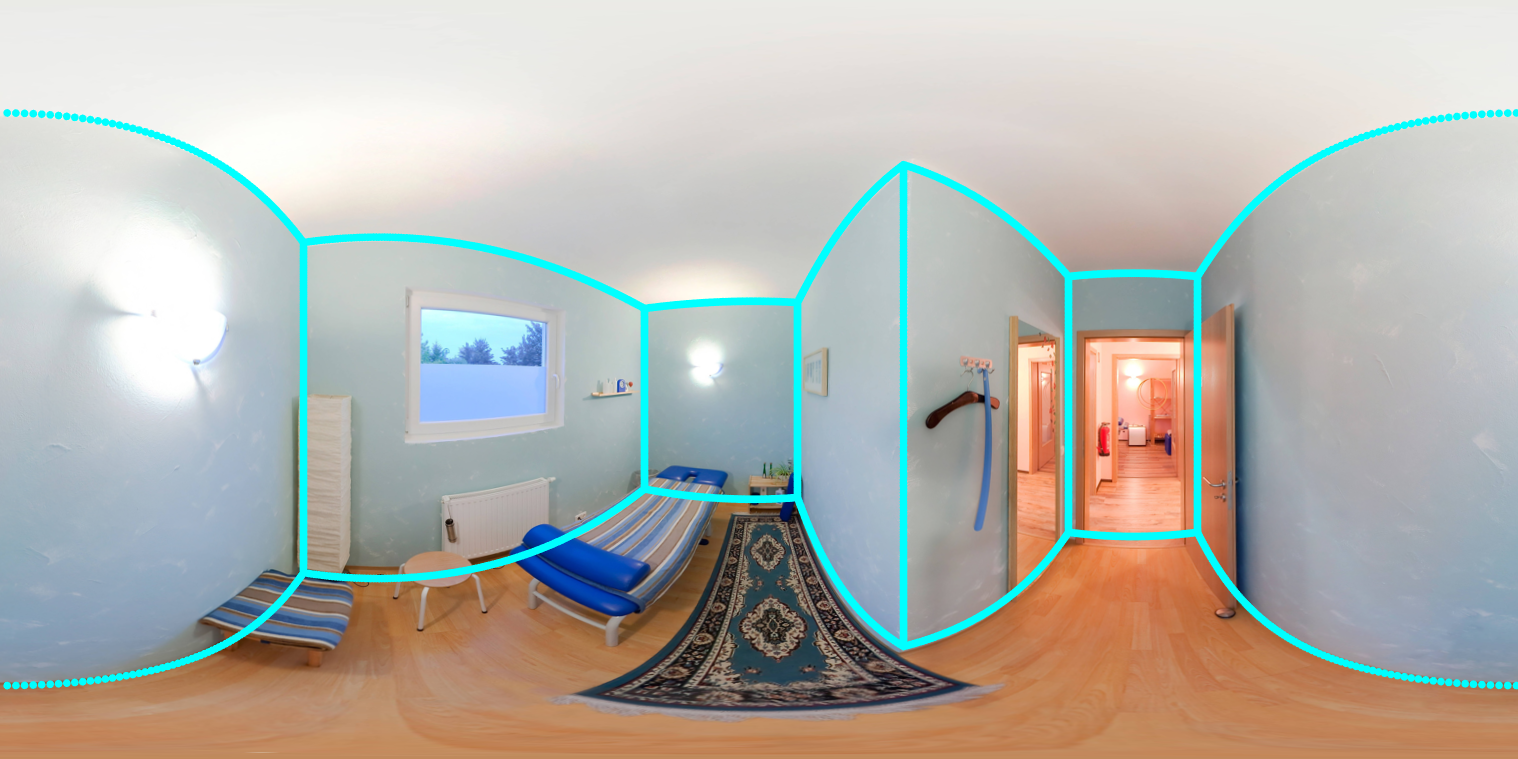}{\label{fig:i4}}}
\end{center}
   \caption{Starting from a single spherical panorama, we exploit the combination of geometry (accurate lines) and deep learning (edge map) to recover the main structure of the room, achieving 3D complex layouts.}
\label{fig:output}
\end{figure}

Layout recovery of indoor scenes is an essential step for a wide variety of computer vision tasks and has recently received great attention from several applications like virtual and augmented reality, scene reconstruction or indoor navigation and SLAM \cite{lukierski2017room}.
%The purpose of recovering indoor layouts from single images has been attempted by numerous researchers, but always with certain limitations. 
Typical constrains are the limited field of view which conducts to obtain open geometries and simple box assumptions considering rooms to have just four walls. The challenge here is to recover closed geometries without strong shape assumptions (Fig.\ref{fig:output}).% avoiding simplifications in a way that can be exploited by both humans and machines, and that is the goal of this paper.

One of the first approaches dealing with indoor layout reconstructions was \cite{Delage2006} which finds floor-wall boundaries by using a Bayesian network model. In contrast, Lee \textit{et al.} \cite{Lee2009} use line segments to generate layout hypotheses evaluating with an Orientation Map, that usually gives problems with clutter since no reasoning about the lines is made. Other works \cite{Hedau2009,Hedau:box,schwing2013box} try to simplify the problem by assuming that the room is a 3D box, which does not match reality in many cases. These proposals rely on Geometric Context, which improves clutter detection compared with Orientation Map but provides worse results at the higher parts of the scenes. More recently, \cite{schwing2012efficient} introduces the concept of integral geometry and pairwise potentials decomposition which results in an efficient structured prediction framework.

\begin{figure*}[t]
\centering
\includegraphics[width=1\textwidth]{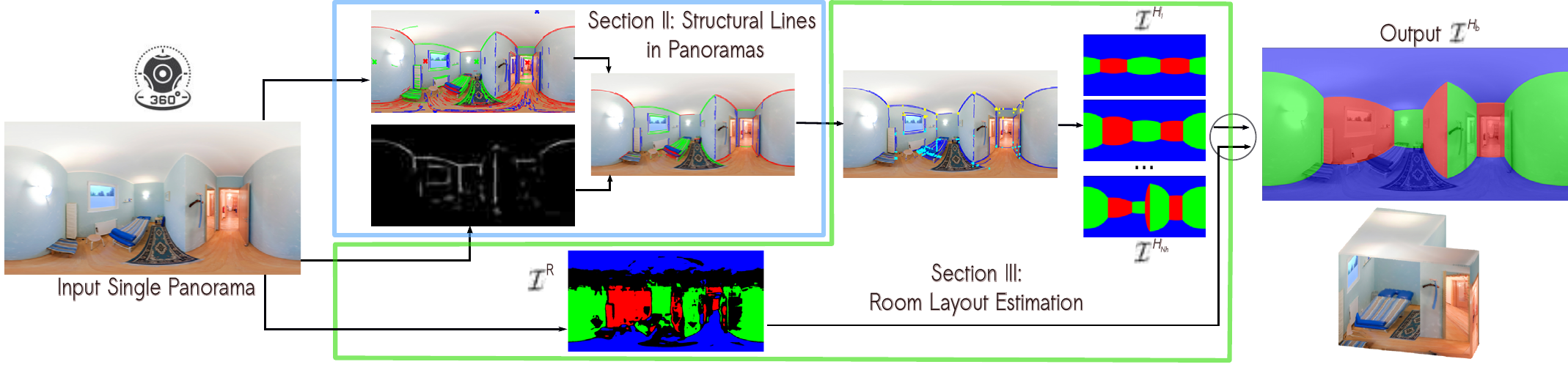}
\caption{\label{fig:pipeline} \textbf{Overview}: From a single panorama, the proposed method combines geometric reasoning (lines and vanishing points) and deep learning (edge map \cite{Mallya:2015}) to generate a pruned set of lines belonging to the main structure of the room from which we extract candidate corners. Layout hypotheses are generated from them and those ones satisfying Manhattan world are evaluated, remaining as the final model the one which fits better with a reference map $\mathcal{I}^{\mathcal{R}}$.}
\end{figure*} 

A crucial limitation of these works is the fact that they use conventional images with limited field of view (FOV). On the one hand, this prevents the reconstruction of the real closed geometry of the whole room. On the other hand, the ceiling does not usually appear, being nevertheless a useful part to detect the main structure of the room as it usually has much less occluding objects than the others.
Therefore, a more recent research direction looks to extend the FOV. Lopez-Nicolas \textit{et al.} in \cite{catad} perform the layout recovery using a catadioptric system. In \cite{perezyus2016peripheral}, layout hypotheses are made combining fisheye images with depth information that provides scale. But the real impact comes with the omnidirectional 360$^{\circ}$ images, which nowadays can be easily obtained with camera arrays, special lenses or automatic image stitching algorithms. This type of images allows to acquire the whole scene at once and hence, it is possible to exploit their wide FOV to generate closed room solutions based on the best consensus distributed around the scene. In \cite{Jia:fisheye}, their method shows the advantages of having a complete scene view over partial views of the same scene \cite{Lee2009}. \textit{PanoContext} \cite{PanoContext} uses panoramas to recover both the layout, which is also assumed as a simple 3D box, and bounding boxes of the main objects inside the room. Similarly, \cite{Pano2cad} provides results not limited to simple box shaped rooms but with the limitation of relying on the output of an object detector.
In \cite{Yang:pano} they treat the problem as a graph with lines and superpixels as nodes, solving it with complex geometric constraints instead.

%However, recent years have seen a growing interest in deep learning. 
On the other hand, in the last years, the research community started to face layout recovery problems with convolutional neuronal networks (CNN) achieving an outstanding success and providing an unprecedented level of data abstraction and pattern recognition that is inspired by neuronal processes. For example, %\cite{yang2016real} uses a CNN to segment the ground plane outperforming traditional methods.
\cite{dasgupta2016delay} provides separate belief maps of the walls, ceiling and floor of the scene.
Alternatively, some works use CNNs to extract the informative structural edges of indoor scenes ignoring those edges from clutter \cite{Mallya:2015,Mallya2}. Instead, in \cite{RoomNet} they predict the location of the room layout corners. Other deep learning works extract an estimation of the depth or/and surface normals from simple RGB images which also produces an interesting outcome for layout estimation \cite{Eigen2015,laina2016deeper}. The main drawback of these CNNs is that they are always focused on traditional images with limited FOV with the consequent limitations we have mentioned before.

In this paper we propose a new entire pipeline which receives as input a 360$^\circ$ full-view panoramic image and returns a closed, 3D reconstruction of the room. Our experimental evaluations in the public databases SUN360 \cite{Xiao2012} and Stanford (2D-3D-S) \cite{2017arXiv170201105A} show that the proposed pipeline (Fig.~\ref{fig:pipeline}) results in high accurate reconstructions outperforming quantitatively ($\downarrow$ pixel error) and qualitatively (greater fidelity to the actual room shapes) the state of the art. The key contributions of the proposed pipeline are the following:
$1)$ The idea of exploiting deep learning combined with geometry to filter non-significant lines. Our proposal allows to work directly with structural lines, and thus structural corners, to create more efficient algorithms that tackle the layout estimation problem with less iterations and more accuracy. %Experiments reveal the advantages of this. 
$2)$ We also propose a new evaluation approach, the Normal Map, alternatively to classical and more recent Maps, and demonstrate to achieve a better performance at hypotheses evaluation step.
$3)$ Finally, we are able to handle flexible closed geometries not limited to 4-wall boxes like other works of the state of the art. This point has a high relevance \textit{e.g.} for using our proposal in a real room-navigation system. Users need to be provided the real space and not just a rectangular simplification of it.

% In this work, we propose a method for layout estimation, where the main novelties are the exploitation of deep learning approaches combined with geometry reasoning, the high flexibility to handle more complex geometries than typical 3D boxes and the utilization of the full 360$^{\circ}$ view that single panoramas provide, allowing us to achieve closed geometries (Fig.~\ref{fig:output}). 
% The pipeline of our method is shown in Fig.~\ref{fig:pipeline}. 
% Starting with a single panorama, first we obtain structural lines by filtering all extracted lines with a deep learning approach. 
% Then we extract good candidate corners from which we generate flexible layout hypotheses and choose the best one by computing a reference map.
% By using only structural lines, we prevent an explosion of the number of layout hypotheses and are able to get accurate final models even in the presence of clutter.
% Experimental evaluation with panoramic images from the public SUN360 \cite{Xiao2012} an Stanford (2D-3D-S) \cite{2017arXiv170201105A} databases of indoor environments shows an improvement with respect to the state of the art and reveals the advantages of combining geometry and deep neural networks in the process. 

%We aim to obtain good candidate corners based on structural lines from which we generate flexible layout hypotheses achieving accurate final models, even in presence of clutter. Structural lines prevent an explosion of the number of layout hypotheses.

\section{Structural lines in panoramas}
\label{sec:cv}
% There are many approaches of line extraction for omnidirectional cameras, such as \cite{bermudez2015automatic} that is able to extract lines for a wide variety of dioptric and catadioptric systems without requiring previous calibration. Other alternative is \cite{catad}, which uses Bazin's Matlab toolbox adapting the equations to hyper-catadioptric system. \textit{PanoContext} \cite{PanoContext} works with panoramas and they split them to a set of perspective images and run the LSD algorithm \cite{von2010lsd} in each one separately and then project the lines back to the panorama. In \cite{oh2012great} they extend the LSD method to deal with panoramas using the great circle arc detector.
In this section we address the initial stage of our proposal, describing how we extract lines and vanishing points (VP) in panoramas, dealing with spherical projection (Section~\ref{sec:lines_vp}). Then we extract the structural lines as a subset of all the lines that are significant for our task as learned from data with a deep learning approach (Section~\ref{mall}).

\subsection{\textbf{Lines and vanishing points estimation}}
\label{sec:lines_vp}
In panoramas, a straight line in the world is projected as an arc segment on a great circle onto the sphere and thus it appears as a curved line segment in the image. For this reason, we represent each line by the normal vector $n_i$ of the 3D projective plane that includes the line itself and the camera center. We adopt the Manhattan World assumption whereby there exist three dominant orthogonal directions. Another particularity of this type of projection is that parallel lines in the world intersect in two antipodal VP whereas in conventional images they do in one single VP. 
In \cite{PanoContext} they split the panorama in order to run a specific algorithm that only works with perspective images, warping then all detected line segments back to the panorama, whereas in \cite{bazin2012globally}, they solve the problem by a branch-and-bound framework associated with a rotation space search.
%We run their algorithm to obtain both lines and VP and it takes $\thicksim$42s per image. We also test the code of \cite{bazin2012globally,bazin20123} obtaining that the computational time amounts to $\thicksim$67s per image working directly on panoramas. 
Here instead, we detect lines and VP by a RANSAC-based algorithm that works directly with panoramas showing entire and unique line segments, avoiding thus duplicate lines coming from different splits and improving the overall efficiency of the method. We achieve really similar results to \cite{PanoContext,bazin2012globally} being also much faster, $\thicksim$8s per image in our proposal, $\thicksim$67s per image with \cite{bazin2012globally} and $\thicksim$42s per image using \cite{PanoContext}.
%We detect lines and vanishing points by a RANSAC-based algorithm that works directly with panoramas, unlike other methods that split the panorama and rectify the images in order to run specific algorithms that only work with perspective images \cite{PanoContext}. 
%Hence, our method shows entire and unique line segments, avoiding duplicate lines coming from different splits and thus improving the overall efficiency of the method, being also faster than \cite{PanoContext} when finding both lines and VP ($\thicksim$8s \textit{vs.} $\thicksim$42s per image).

%%vanishing points (VPs) in the image aligned with three dominant directions in the world. 

% \begin{figure*}
% \begin{center}
%    \subfloat{\includegraphics[width=0.32\linewidth]{pano_vp_lines_5.png}}\hspace*{0.01\linewidth}
%    \subfloat{\includegraphics[width=0.32\linewidth]{panoEdges_5.png}} \hspace*{0.01\linewidth}
%    \subfloat{\includegraphics[width=0.32\linewidth]{pano_vp_lines_5_2.png}}
% \end{center}
%    %\caption{Combined information}
%    \caption{\textbf{Left}: lines and vanishing points extraction with geometric reasoning. \textbf{Center}: Edge Map obtained through the FCN from Mallya \textit{et al.} \cite{Mallya:2015} . \textbf{Right}: Resulting structural lines after combining geometry an deep learning showing a large reduction of the number of lines, while those more significant for the main structure remain.}
%    \label{fig:edges_lines}
% \end{figure*}

\begin{figure}
\begin{center}
   \subfloat{\includegraphics[width=0.99\linewidth]{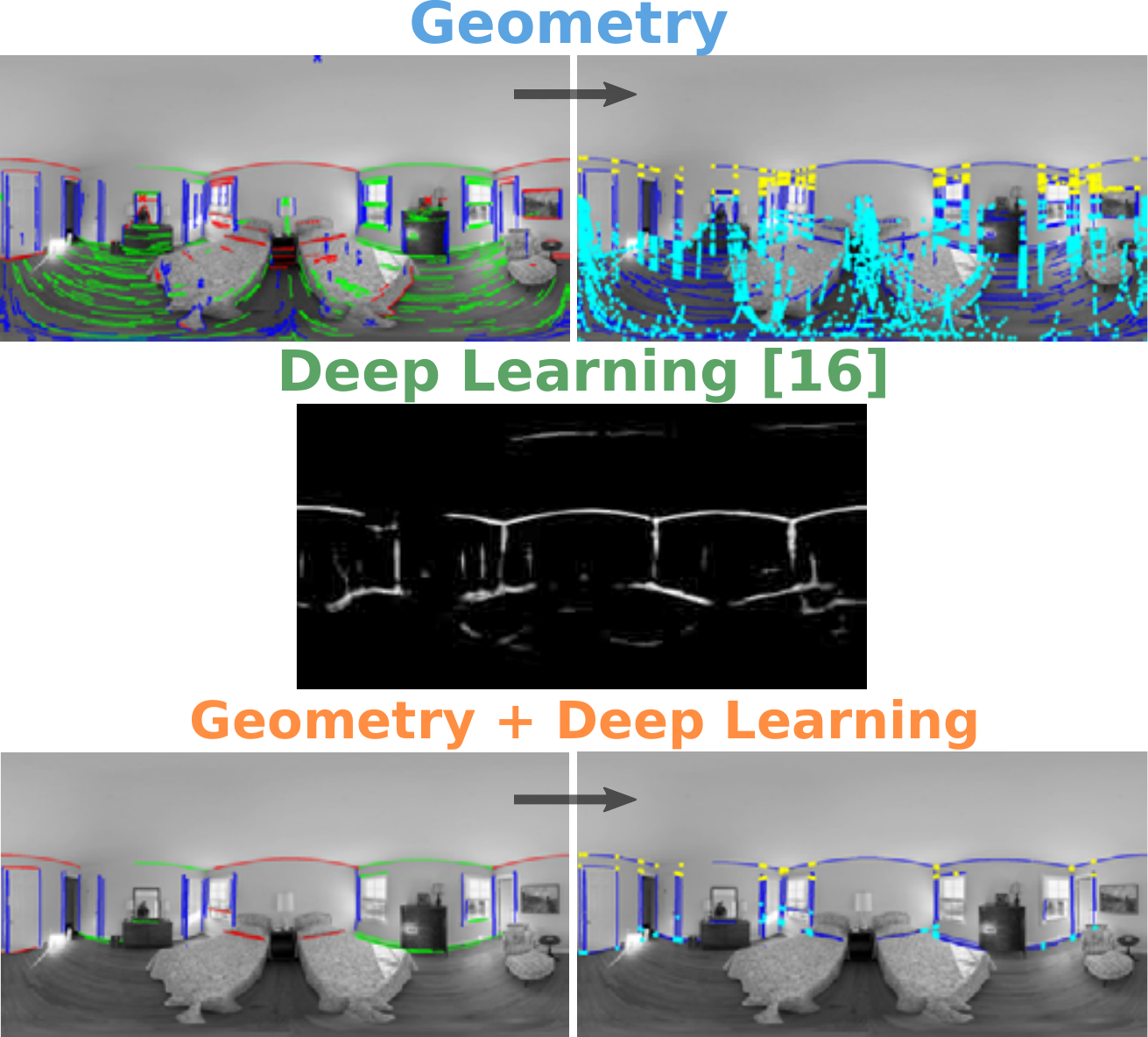}}
\end{center}
   %\caption{Combined information}
   \caption{\textbf{Top}: Oriented lines and corners extracted just with geometric reasoning. \textbf{Center}: Edge Map obtained through \cite{Mallya:2015}. \textbf{Bottom}: Resulting structural lines and corners after combining geometry an deep learning. A large reduction of them is shown, while those more significant for the main structure remain. Corners become good candidates for the hypotheses generation.}
   \label{fig:edges_lines}
\end{figure}

\begin{figure}
\begin{center}
   \subfloat{\includegraphics[width=0.99\linewidth]{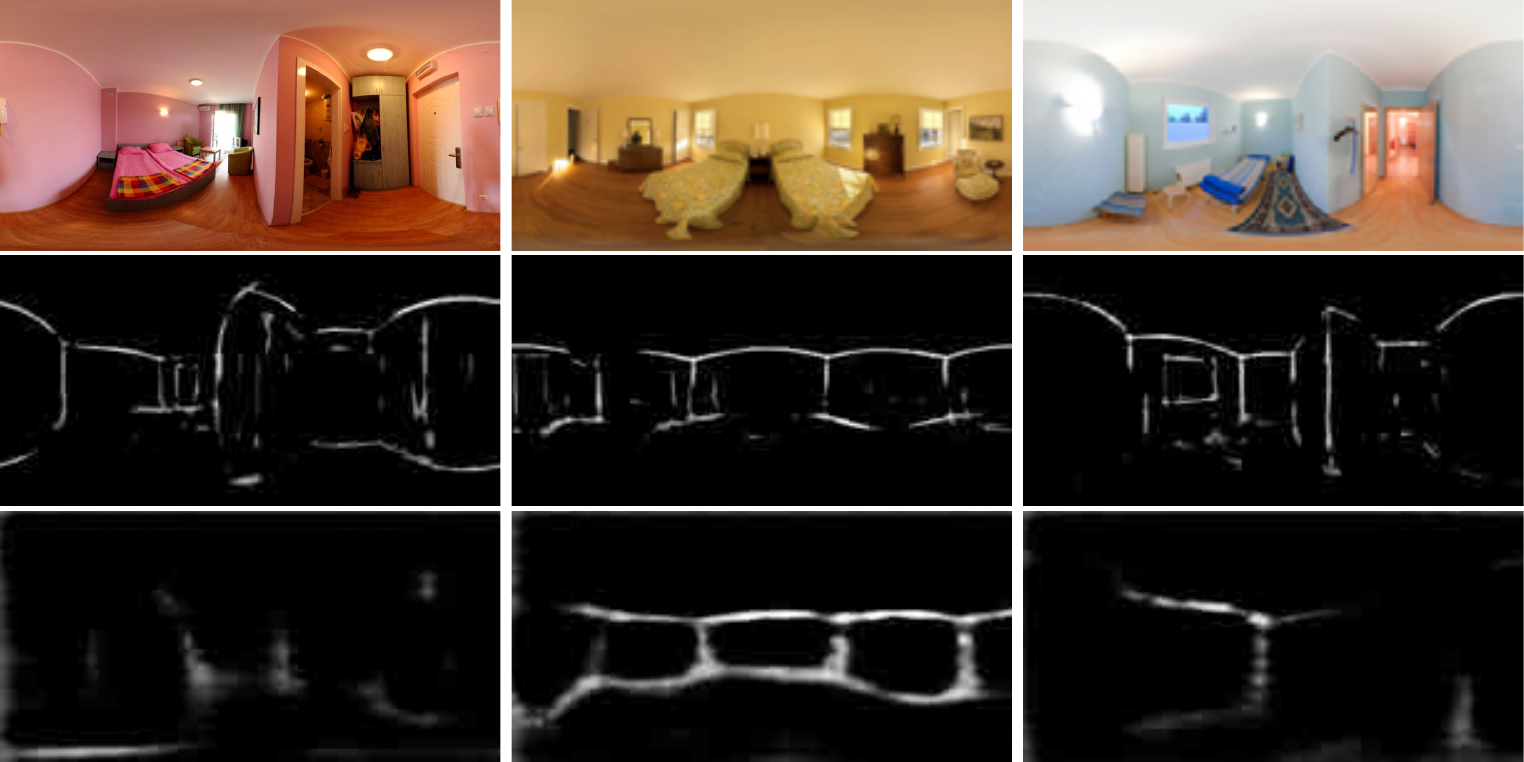}}
\end{center}
   %\caption{Combined information}
   \caption{Comparison of edge maps obtained applying \cite{Mallya:2015} through the proposed discretization of the sphere (\textbf{Center}) and directly on the panorama (\textbf{Bottom}).}
   \label{fig:edgemaps}
\end{figure}

First, we run a Canny edge detector on the panorama and cluster contiguous edge points in edge groups. % considering them as candidate lines of the image. 
Each point of the edge group $i$ is projected into the 3D space as a spatial ray $r_{ij}$, $\forall j=\left\{1..N_{pts}\right\}$.
%Each group $i$ projects spatial rays in the 3D space $r_{ij}$ for each point of the image, i.e. $\forall j=\left\{1..N_{pts}\right\}$.
Iteratively, two points of each group are randomly selected ($r_{i1}$, $r_{i2}$) %and projected into the 3D space as spatial rays ($r_{i1}$, $r_{i2}) \in r_{ij}$ with $j={1..N_{pts}}$. 
and thus we get a possible normal direction for the edge group $n_{i} = (r_{i1} \times r_{i2})$.
The number of \textit{inliers} is evaluated, \textit{i.e.} how many rays fulfill the condition of perpendicularity with the normal under an angular threshold of $\pm 0.5^{\circ}$, $|arccos(n_i\cdot r_{ij})-\frac{\pi}{2}|\leq \theta_{th}$. 
After a certain number of iterations the process outputs, for each edge group, the model leading to the highest number of inliers giving the $n_i$ that fits the line best. %Edge groups with few points are discarded. 
%Finally, we get a set of well-defined lines used to obtain VP directions ($vp_k$) applying a RANSAC algorithm, considering $vp_k = n_a \times n_b$ where $n_a$ and $n_b$ are the normal vectors of two world parallel lines. 

We obtain the three orthogonal VP directions ($vp_k$) with another RANSAC algorithm, considering $vp_k = n_a \times n_b$ where $n_a$ and $n_b$ are the normal vectors of two world parallel lines.
Eventually we select the three VPs ($vp_x, vp_y, vp_z$) that have the most number of inlier lines, exploiting that normal vectors $n_i$ must be orthogonal to the main directions $|arccos(n_{i}\cdot vp_{k})-\frac{\pi}{2}|\leq \theta_{th}$ where $k={x,y,z}$. 
%The line segments are classified according to the Manhattan directions (VPs). To do that we exploit that normal vectors $n_i$ must be orthogonal to the main directions $|arccos(n_{i}\cdot vp_{k})-\frac{\pi}{2}|\leq \theta_{th}$ where $k={x,y,z}$. 
Inlier lines are classified according to the VP, whereas the other lines are discarded (those whose normals are not perpendicular to any of the main directions).
%Lines are classified according to the VP they are inliers to, and lines whose normals are not perpendicular to neither of the directions are discarded.
The lines with the same Manhattan direction are shown in identical color in Fig. \ref{fig:edges_lines} (top-left).
Once VP are computed, we rotate the panorama in a way that it is pointed perpendicularly to one of the room walls.
%Although there are methods that guarantee a global optimum \cite{bazin2012globally}, RANSAC methods are faster and robust.%we also achieve good enough results.

\subsection{\textbf{Structural lines introducing deep learning}} 
\label{mall}
The main piece of information we use to create layout hypotheses are lines. However, in cluttered scenes is very difficult to know whether they come from actual wall intersections or from other elements of the scene. Proceeding with all the lines leads to an intractable number of hypotheses. In order to tackle this problem, we propose to evaluate the extracted lines on the panoramic image introducing deep learning. %? 
CNNs have been successfully applied to extract complex features such as corners \cite{RoomNet} or structural edges \cite{Mallya:2015}. However, they have not been trained to deal with omnidirectional images and then, they are very inaccurate when used directly on panoramas. %results of applying them directly on the panoramic images are very inaccurate. 
Besides that, it does not exist any dataset collecting panoramic images with enough amount and variety of labeled data required to train a deep neural network. Thus, we do not directly train an end-to-end CNN and decide, instead, to adapt an existing CNN to our image geometry.
Here, we adapt the Fully Convolutional Network (FCN) proposed by Mallya and Lazebnik \cite{Mallya:2015}. This network was trained to estimate probability maps representing the room edges of the projected 3D box that fits the room better, even in the presence of clutter and occlusions. 
Our proposal is to combine such rough yet meaningful information with more accurate geometric cues such as lines. 

%to select structural lines demonstrating the advantages of combining the accurate lines extracted by geometry with deep learning feature abstraction by adapting an existing CNN.%Thus, we propose the selection of structural lines with deep learning techniques by adapting an existing method, instead of directly train an end to end CNN.
%this is also the reason why we do not directly train an end-to-end CNN and decide, instead, to demonstrate the advantages of combining geometry precision with deep learning by adapting existing methods.

To apply the FCN, we split the panoramas into a set of overlapping perspective images with a FOV similar to conventional images ($\thicksim$70$^{\circ}$) and planar projection. We run the algorithms in each of them separately to obtain local results %for the original 360$^{\circ}$ image 
and finally stitch them all back to the panorama as in \cite{Xiao2012,PanoContext,Pano2cad}. 
For the discretization of the sphere, instead of selecting 
%A first attempt to define the set of virtual perspective images compounding the panorama would be to select 
spherical coordinates from uniform distributions $\theta\in(-\pi/2,\pi/2)$ and $\phi\in(-\pi,\pi)$ (which is not adequate since the density increases as we get closer to the poles), we use an algorithm based on the golden section spiral \cite{gonzalez2010measurement}. %\footnote{\textbf{Points on a sphere} \url{http://www.softimageblog.com/archives/115}}.
For any given number of points, it results in an evenly distribution with bins covering areas of similar size equally distant from their closest neighbor. We experimentally choose 60 points, \textit{i.e.} 60 perspective images.

To improve the edge maps, we avoid noise by removing low probability pixel values below a certain threshold (0.2 out of 1).  
When the virtual perspective images are stitched back to the panorama, there are some overlapping regions that we solve by choosing the maximum value of probability to not lose information. In Fig \ref{fig:edgemaps} we show that the accuracy of the edge map substantially improves when we split the panorama, specially in those cases where the result of applying directly the FCN on the panorama is completely uninformative (first and third columns).
%Fig. \ref{fig:edges_lines} (Center) shows an example of informative edges detection on a panoramic image.
Once we have the edge map of the panorama given by the FCN \cite{Mallya:2015}, we give each extracted line a score calculated as the sum of the corresponding probability values to the pixels it occupies in the edge map. In this way, we remove those lines whose score is below a certain threshold (the 10\% of their length), while the others are classified as structural lines. 
An example of this process can be observed in Fig. \ref{fig:edges_lines}. It shows clearly the advantage of merging both approaches, where %which allows us to work directly with the more significant lines of the structure of the room, ignoring 
those lines that belong to clutter such as those from the parquet, the tables and even many windows, pictures and doors have been removed, but most relevant lines to recover the structure of the room remain for further stages. %After carrying out this merging of information, 
With this operation the number of lines may be reduced to one-third or even a quarter depending on the scene.

\section{Room layout estimation}
\label{sec:layouts}
Our goal is to extract the main structure of an indoor environment \textit{i.e.} the distribution of floor, ceiling and walls, abstracting all objects within rooms. For this purpose we have developed a method to generate layout hypotheses from corners found with the already filtered significant lines. Our algorithm is divided in three stages:

% \begin{figure} 
% \begin{center}
%    %\subfloat{\includegraphics[width=1\linewidth]{FCN.pdf}{\label{fig:corners1}}}
%    \subfloat{\includegraphics[width=0.5\linewidth]{corners_MIL.png}{\label{fig:corners1}}} \hspace*{0.003\linewidth}
%       \subfloat{\includegraphics[width=0.5\linewidth]{all_corners_5.png}{\label{fig:corners2}}}
% \end{center}
%    \caption{\textbf{Candidate corners} from both ceiling (yellow) and floor (cyan). \textbf{Left}: using only geometric reasoning. \textbf{Right}: from structural lines obtained thanks to geometry and deep learning combination. Corners became good candidates for the hypotheses generation.}
%    \label{fig:corners}
% \end{figure}

\begin{figure*} [h]
\centering
\includegraphics[width=1\textwidth]{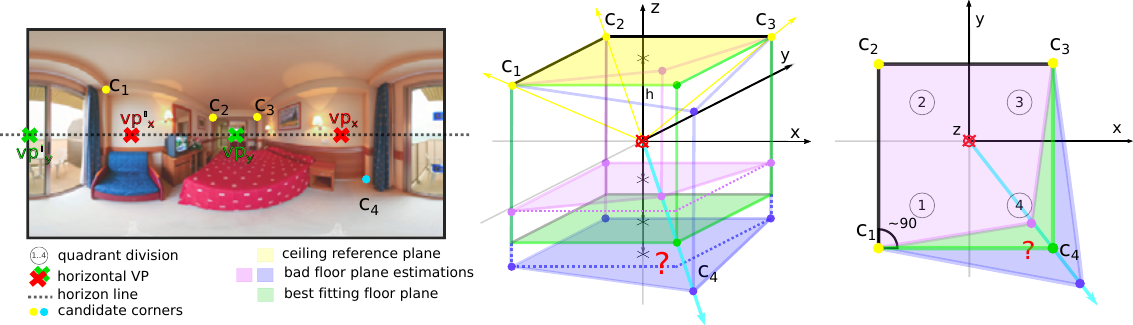}
\caption{\label{fig:floor_plane}\textbf{Room height}. We take advantage of the ceiling-floor symmetry to estimate the distance between both planes. We look for the solution that makes the projection of the floor corner such that the walls connected by the corners are as perpendicular as possible (Manhattan world assumption).}%\textbf{Room height decision}: The best fitting solution fulfilling the Manhattan assumption provide us an estimated room height for our layout hypothesis.}
\end{figure*} 

\subsection{\textbf{Candidate corners extraction}}
Our layout generation process is based on corners, \textit{i.e.} structural intersections between two walls and ceiling or floor. 
In a Manhattan World, two line segments are enough to define a corner, so we intersect all the significant lines in different directions $(x,y,z)$ among themselves in pairs as long as they do not cross each other. 
The direction vector of the corner point is computed with a cross product of the lines intersecting in that corner, $ c_ {ac} = (n_a \times n_c)$. %Each corner has an associated score given by the scores sum of the lines that have generated this corner. Then, those with a score lower than a certain threshold are removed, facilitating the subsequent generation of hypotheses. 
The previous selection process of structural lines with learning makes these extracted corners already good candidates. 
Fig. \ref{fig:edges_lines} shows the large difference between obtaining corners with first line extraction (top) and with structural lines (bottom). By removing non-structural lines, the number of corners extracted is vastly reduced, yet the important ones remain detected. This reduction makes further stages of the method faster and more efficient, but also improves the reliability of the results since most corner candidates coming from clutter and irrelevant structures are not considered.

Panoramic images have the advantage of providing a full view of the room, allowing us to look around, up and down in the scene. Unlike conventional images where the ceiling and some walls use to be out of the FOV. Taking this into account, we carry out a classification of the detected corners following two criteria (See Fig.\ref{fig:floor_plane}):
\begin{enumerate}
\item Their position along the $z$ axis:
Corners detected below the horizon line $l_{H} (-z)$ in the image are considered as floor corner candidates and those detected above the $l_{H} (+z)$ as ceiling corner candidates. %(The $l_{H}$ goes through the horizontal VP). 
\item Their position in the $XY$-plane: 
Since the camera is inside the room, we divide the scene into four quadrants around the center of the camera with the horizontal VPs as quadrant dividers, $\mathcal{Q} = \{q_{1},q_{2},q_{3},q_{4}\}$. Hence, \textit{e.g.} to determine when a corner belongs to the fourth quadrant: $ c \in q_{4} \Longleftrightarrow c_{x}\in \mathbb{R}^{+} \wedge c_{y}\in \mathbb{R}^{-} $.
\end{enumerate}
%Panoramic images have the advantage of providing a full view of the room, allowing us to always observe not only the floor as in conventional images, but also the ceiling. While we do not have the 3D coordinates of the corners but just their image projection, we assume that the candidate corners from each hemisphere (above and below the horizon line in the image) intersect in a single ceiling and floor plane respectively. The normal direction of both planes is the vertical Manhattan direction. In this work, we do not make pairing between ceiling and floor corners since they do not usually appear both in the image due to occlusions. Instead, when generating layout hypotheses, we will obtain the ceiling-floor height relationship that will allow us to implicitly obtain the complementary one by symmetry. 
\begin{figure} [h]
\centering
\includegraphics[width=0.5\textwidth]{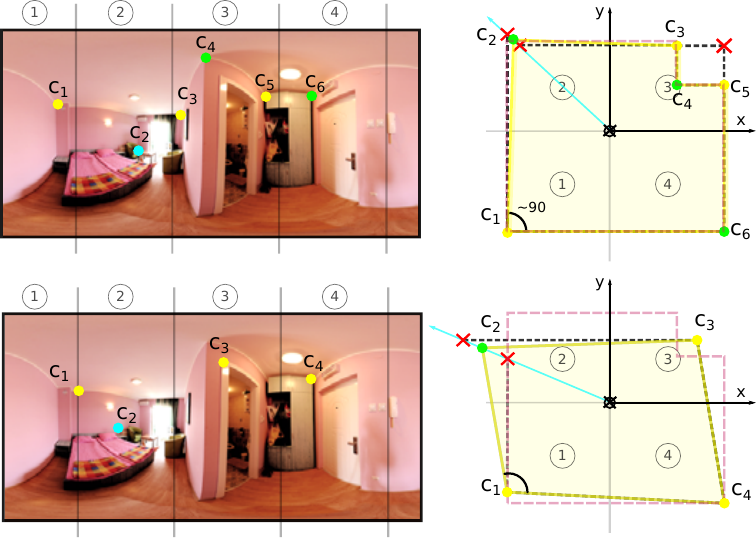}
\caption{\label{fig:hypot_6} \textbf{Layout hypothesis generation}: We show two examples of layout hypotheses generation. 
The first example corresponds to a valid hypothesis whereas the second one presents a non-valid discarded hypothesis.}
\end{figure} 

Manhattan World rooms always have an even number of walls, and the number of corners in each quadrant will be an odd number so, this quadrant division allows to easily know some additional information to sample corners. For example, the simplest layout will include just one corner in each quadrant while more complex layouts will have three or even five corners in some of their quadrants.
%After evaluating this last disposition we found that it is very useful since we know that, \textit{e.g.}, in rooms with four walls there is always a corner in each quadrant or, in more complex rooms, \textit{e.g.} rooms with six walls, there is always a corner in each quadrant except in one of them where there will be three corners.

\subsection{\textbf{Layout hypotheses generation}} 
Many works simplify the layout generation problem by assuming that the room is a simple box of four walls, sometimes because of lack of information due to the use of conventional images with smaller FOV \cite{Hedau2009,Hedau:box,schwing2013box}, or just to subtract complexity to the problem \cite{PanoContext}. Here, we face more complex designs which will be faithful to the actual shapes of the rooms, introducing the possibility of estimating in-between hidden corners when required, \textit{i.e.} when they are occluded by clutter or due to scene non convexity. 
We generate layout hypotheses by means of an iterative method that attempts to join consecutive corners with alternatively oriented walls following Manhattan assumption. 
%We initially proceed to make a clear distribution of the scene, we rotate panoramas in a way that it is pointed perpendicularly to one of the room walls and, since camera is situated inside the sphere, we divide the scene into four quadrants around the center of the camera with the horizontal VPs as quadrant dividers, $\mathcal{Q} = \{q_{1},q_{2},q_{3},q_{4}\}$. Hence, \textit{e.g.} to determine when a corner belongs to the fourth quadrant: $ c \in q_{4} \Longleftrightarrow c_{x}\in \mathbb{R}^{+} \wedge c_{y}\in \mathbb{R}^{-} $ %c_{x}\in[vpx'_{x},vpy'_{x}] \wedge c_{y}\in[vpx'_{y},vpy'_{y}]$. See Fig.\ref{fig:floor_plane}.
%In order to reduce the number of iterations needed, we initially proceed to make a clear distribution of the scene, dividing it into four quadrants around the center of the camera with the horizontal VPs as quadrant dividers. 
%Knowing this distribution 
%After evaluating this disposition we found that it is very useful since we know that, \textit{e.g.}, in rooms with four walls there is always a corner in each quadrant or, in more complex rooms, \textit{e.g.} rooms with six walls, there is always a corner in each quadrant except in one of them where there will be three corners.

Our algorithm randomly generates at each iteration initial groups of corners, $\mathcal{G}_{c}$, which are ordered clockwise in the $XY$-plane. There is a relation between the number of corners randomly selected $N_{\mathcal{G}_{c}}$ and the maximum number of walls $N^{max}_{W}$ that our algorithm is able to solve with them, $N^{max}_{W} = 2 (N_{\mathcal{G}_{c}} - 1)$. In this way, we can adjust the complexity of the layouts just giving more or less freedom to the random function that selects the initial corners. This relation means that \textit{e.g.} we can draw layouts with six walls from a minimum of four corners allowing the algorithm to introduce two new corners that maybe were not visible in the image.
For this initial selection, we establish a minimum requirement for which there must be corners in at least three quadrants $\subseteq\mathcal{Q}$, thus, the corner in the remaining quadrant can be estimated assuming closed Manhattan layouts, and there must be at least one corner of each hemisphere, \textit{e.g.} $\mathcal{G}_{c}=\{c^{ceiling}_{q_{2}},c^{floor}_{q_{3}},c^{ceiling}_{q_{4}}\}$. 
% Our algorithm for layout hypotheses generation starts with a pseudo-random selection of possible corners subject to the next conditions:
% \begin{itemize}
% \item 
% Initial groups of three, four or five corners are randomly generated at each iteration and ordered clockwise.
% \item 
% There must be corners in at least three of the four quadrants. Thus, the corner in the fourth quadrant can be estimated assuming closed Manhattan layouts.
% \item 
% In the selected group there must be at least one corner of each hemisphere, \textit{i.e.} at least one ceiling and one floor corner.
% \end{itemize}
This last condition allows us to estimate the height of the room, \textit{i.e.} the relative distance of the camera to the ceiling and floor planes. 
We proceed with the geometric reasoning in 2D as in the right side of the Fig.\ref{fig:floor_plane}, with a top view of the scene. While we do not have the 3D coordinates of the corners but just their direction vector (ray), we assume that all the candidate corners from each hemisphere intersect in a single ceiling and floor plane respectively. The vertical Manhattan direction is the normal direction of both planes (Ceiling-floor symmetry).

An example of hypotheses generation performing this operation is shown in Fig. \ref{fig:floor_plane}. 
The corners above the $l_{H}$ ($c_1,c_2$ and $c_3$) belong to the ceiling, so we intersect their rays (yellow) into a reference ceiling plane in a way that the walls connecting them can be obtained in 2D assuming Manhattan world. The corner below the $l_{H}$ ($c_4$) belongs to the floor but, although we know that it is parallel to the ceiling plane, the distance between them is a priori unknown. We use the Manhattan world requirement to estimate the floor position along its ray (cyan), choosing the one that makes the projection of the point such that the walls connected by the corner are as perpendicular as possible. %The hypotheses are up to scale. %, unless previous knowledge is provided (\textit{e.g.} in \cite{PanoContext,Pano2cad} they assume the camera center is at 1.7m). 

% As the Fig. \ref{fig:floor_plane} shows, the candidate corners are projected in the $x-y$ plane of the sphere model and are ordered clockwise. The corners above the horizon line are projected as a point ($c_1,c_2$ and $c_3$) on the ceiling reference plane, while the corner below the horizon line is projected as a ray ($c_4$) along which we find the best fitting floor plane for the design model we are looking for. Then layouts are generated by joining corners in order with Manhattan-oriented walls whenever possible. Whenever the set of corners withdrawn cannot generate layout hypotheses with alternatively oriented walls satisfying Manhattan assumption, a new set of corners is selected. Without loss of generality, like in previous works ~\cite{PanoContext, Pano2cad}, we assume that the camera center is placed at a typical height (\textit{e.g.} 1.7 meters), which allows us to compute the floor plane and therefore the position of the corners in 3D. Our method finds the ceiling height so that the 3D position of the corners would produce the best Manhattan layout. Due to  the ceiling-floor symmetry, either a point in the ceiling-wall boundary or in the floor-wall boundary is sufficient to specify both. 

In Fig. \ref{fig:hypot_6} two more complex examples of layout hypotheses generation are shown. In the first example we present a valid layout hypothesis where an initial random group of candidate corners is selected $\mathcal{G}_{c} = \{c_1,c_2,c_3,c_5\}$. This means that the algorithm will be able to solve a layout hypothesis with $N^{max}_{W}=6$. A joining corner process starts from $c_1$ finding then a floor spatial ray. In order to find the optimal corner position along this ray, the algorithm finds possibilities with its nearest corners and draws an intermediate solution, $c_2$. In the third quadrant, taking into account the direction ($x-y$) from previous unions, our algorithm selects the best solution for $c_4$ by choosing the one which produces alternatively oriented consecutive walls. In the empty quadrant, Manhattan walls from nearest corners give $c_6$. For each union the Manhattan assumption is checked with a certain threshold ($90^{\circ} \pm 5^{\circ} $). In the second example we show a non-valid layout hypothesis. Following the same idea, initial random corners are selected ($c_1,c_2,c_3,c_4$) and orderly joined conforming in this case a non-Manhattan layout, so it is rejected as hypothesis. When we get the corner floor position along its direction vector, we can obtain the distance between ceiling and floor planes that verifies the ray equation (This is illustrated in the \textbf{video attachment}).
% \begin{figure} [h]
% \begin{center}
%    \subfloat{\includegraphics[width=0.5\linewidth]{BestCorners_normals_5_b.png}}\hspace*{0.01\linewidth}
%    \subfloat{\includegraphics[width=0.5\linewidth]{5.png}} \hspace*{0.01\linewidth}
%    %\subfloat{\includegraphics[width=0.32\linewidth]{corners_5.png}}
% \end{center}
%    %\caption{Combined information}
%    \caption{\textbf{Left}: Reference Map obtained trough \cite{Eigen2015} $\mathcal{I}^{NM}$ for hypotheses evaluation. \textbf{Right}: Map of normals generated from the best hypothesis at the evaluation step.}
%    \label{fig:normal_lines}
% \end{figure}

\begin{figure*} [h]
\begin{center}
\subfloat[Layout hypothesis ($\mathcal{I}^{H_i}$)]{\includegraphics[width=0.195\linewidth]{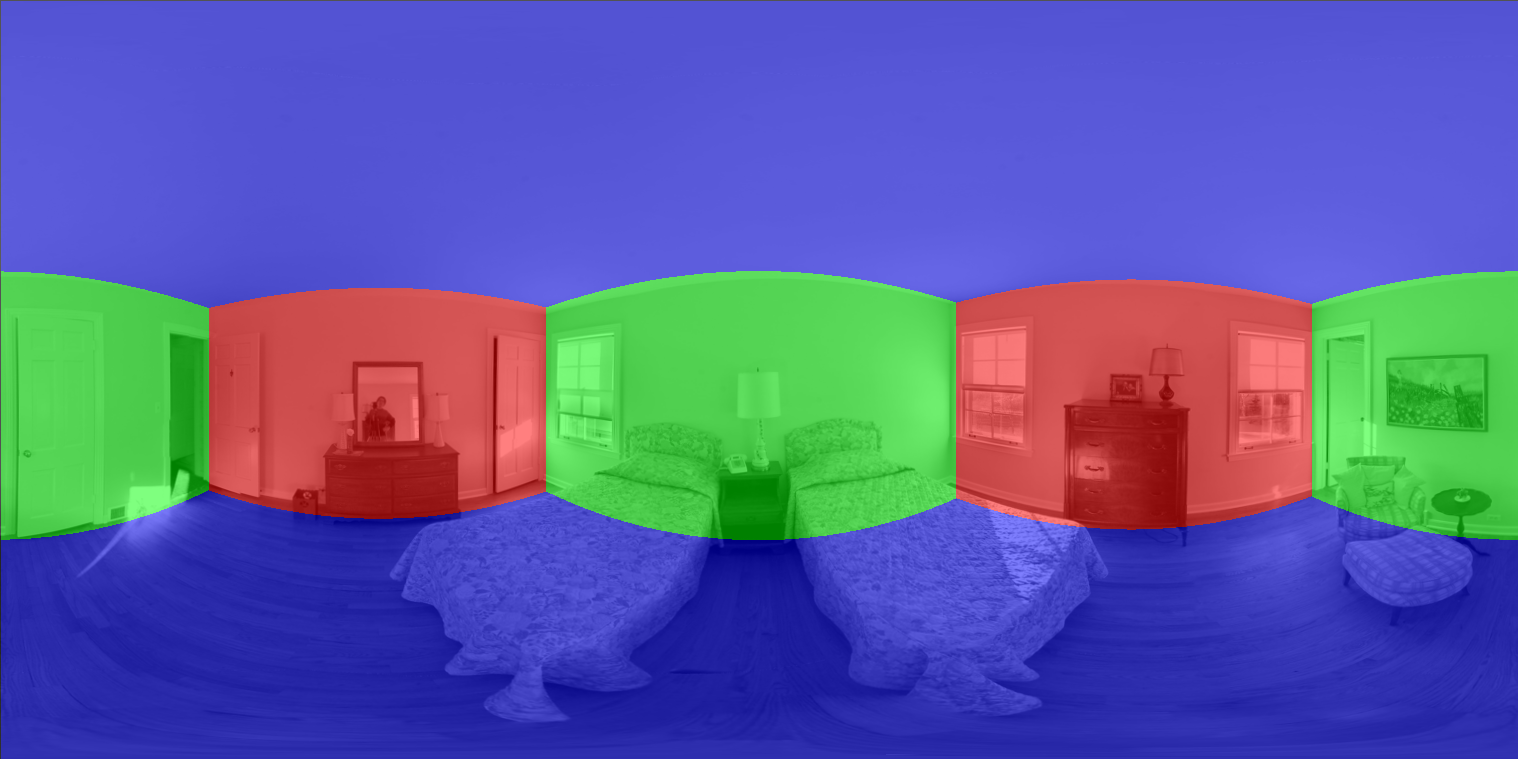}} \hspace*{0.005\linewidth}
   \subfloat[Normals Map ($\mathcal{I}^{NM}$)]{\includegraphics[width=0.195\linewidth]{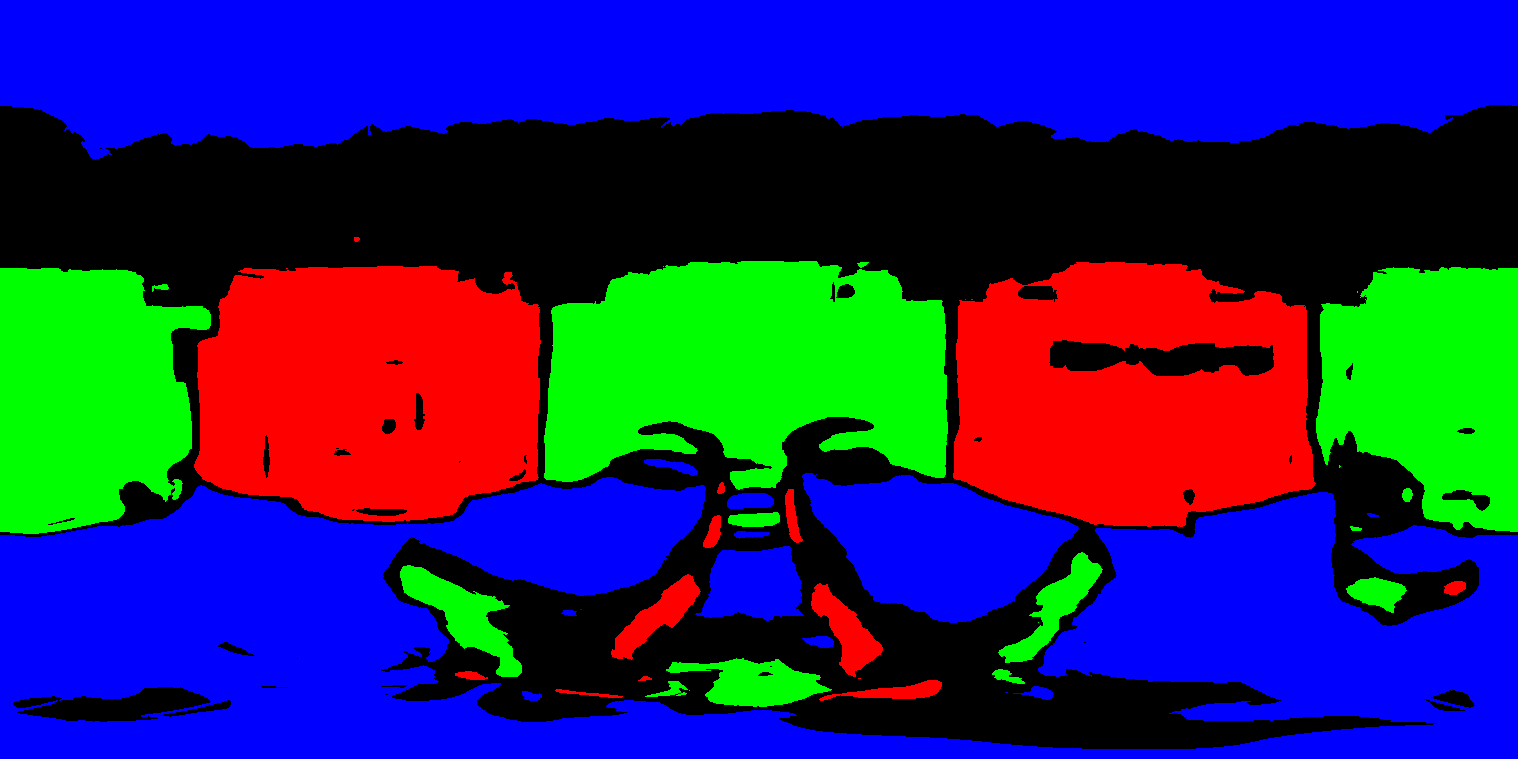}} \hspace*{0.005\linewidth}
   \subfloat[Orientation Map ($\mathcal{I}^{OM}$)]{\includegraphics[width=0.195\linewidth]{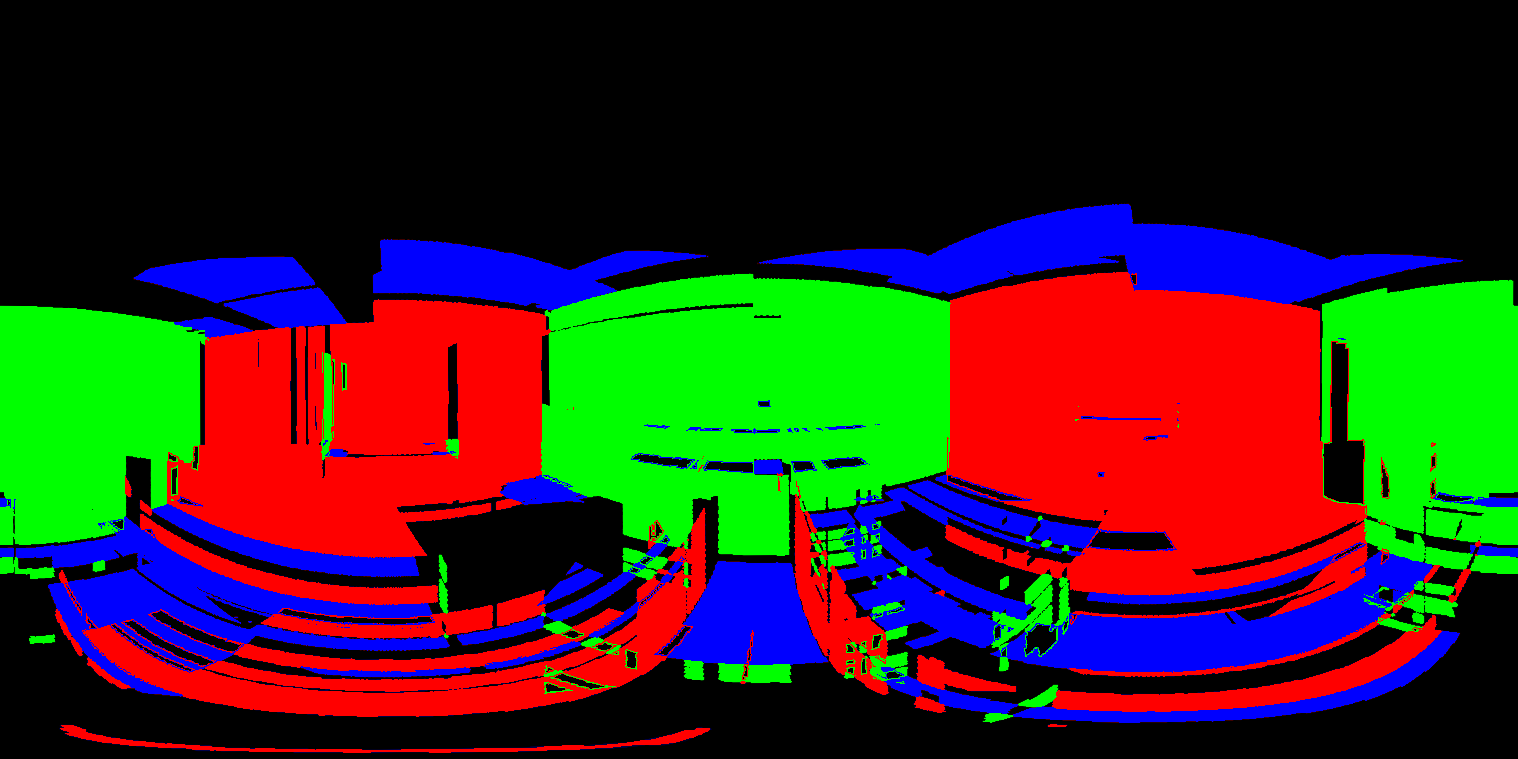}} \hspace*{0.004\linewidth}
   \subfloat[Geometric Context ($\mathcal{I}^{GC}$)]{\includegraphics[width=0.2015\linewidth]{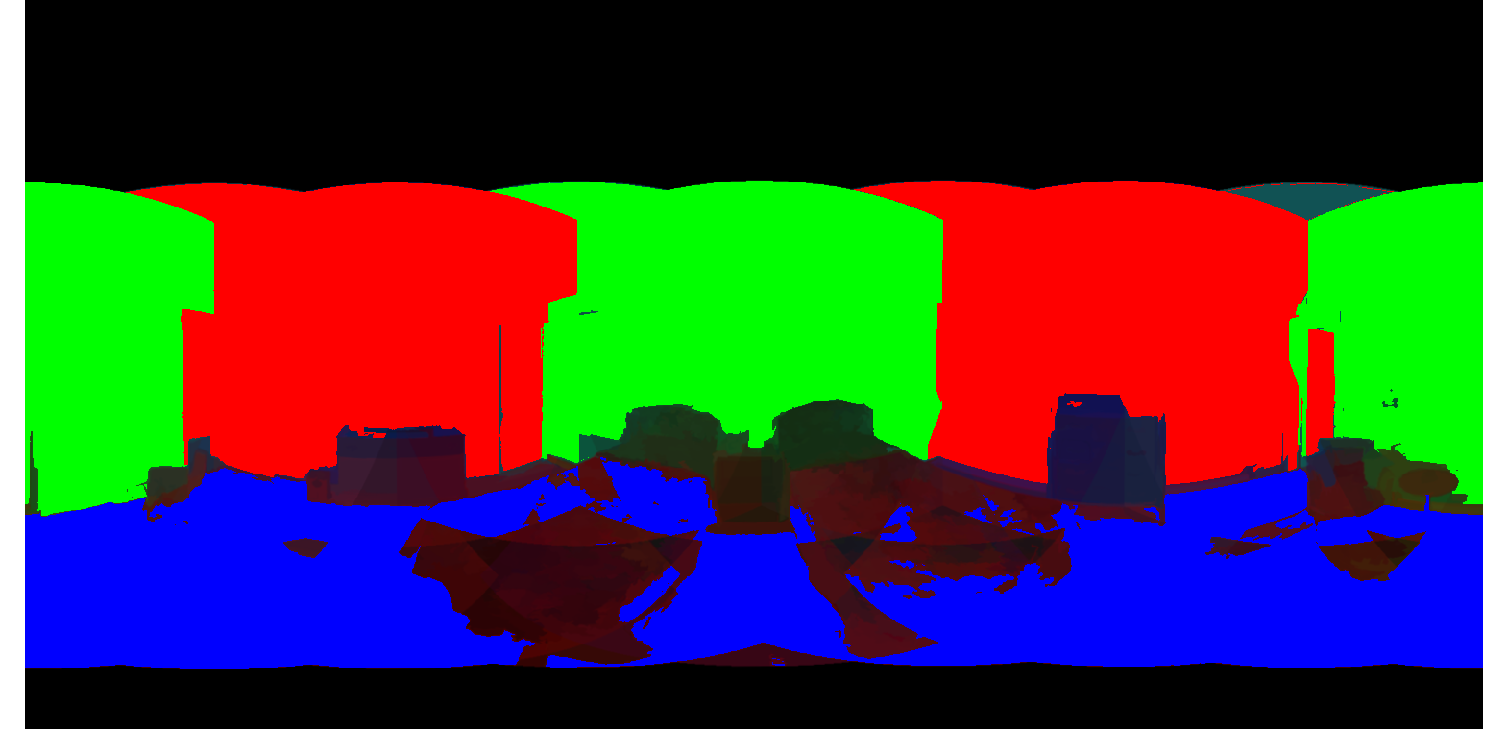}} \hspace*{0.004\linewidth}
   \subfloat[Merge Map ($\mathcal{I}^{MM}$)]{\includegraphics[width=0.195\linewidth]{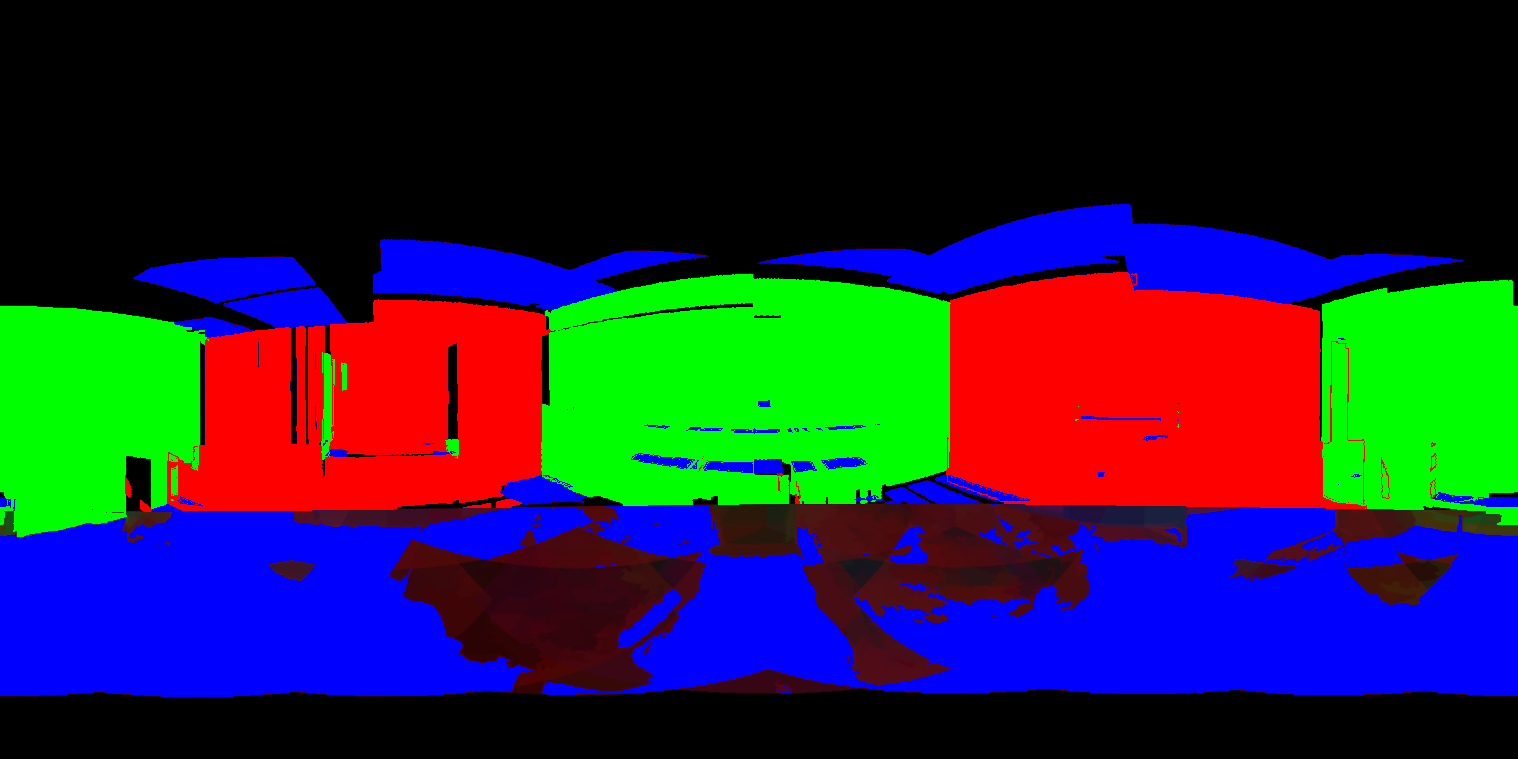}} %\hspace*{0.01\linewidth}
   %\hspace*{0.01\linewidth}
    %\hspace*{0.01\linewidth}
   %\subfloat{\includegraphics[width=0.32\linewidth]{corners_5.png}}
\end{center}
   %\caption{Combined information}
   \caption{(a) Example of labeled image generated from layout hypotheses. (b)-(e) Visual representation of how each of the \textit{reference maps} $\mathcal{I}^{\mathcal{R}}$, looks like.}
   \label{fig:normal_lines}
\end{figure*}

\subsection{\textbf{Layout hypotheses evaluation}} 
\label{hev}
In the hypotheses generation stage we obtain a certain number of layout hypotheses ($N_h$).
In the evaluation process we determine which one is the best, and therefore, the final result.
For each hypothesis $H_i$, we generate a labeled image $\mathcal{I}^{H_i}$, in which each pixel encodes the orientation of the surface (e.g. wall in $x$, wall in $y$ or floor/ceiling in $z$).
%For each i$\mathcal{I}^{H} = \lbrace \mathcal{I}^{H}_{1}..\mathcal{I}^{H}_{N_h}\rbrace$
In Fig. \ref{fig:normal_lines}(a) there is an example of a labeled map $\mathcal{I}^{H_i}$ where each label has different color.
Then, we evaluate the hypotheses fitting to a \textit{reference map} $\mathcal{I}^R$ that roughly encodes the orientation of the pixels, and can be obtained from several methods.
We compute the ratio of pixels that are equally oriented in $\mathcal{I}^R$ and $\mathcal{I}^{H_i}$ over the total size of the image, that we call Equally Oriented Pixel ratio ($EOP$):% Eq. \ref{acc_e}.

%\begin{equation}
\begin{align*}
EOP\left(\mathcal{I}^{H_i},\mathcal{I}^R\right) = \frac{1}{M \cdot N} \sum_{x,y,z}^{P} \sum_{i,j}^{M,N} \mathcal{I}^{H_i}\& \mathcal{I}^{\mathcal{R}} ,
%\label{acc_e}
\end{align*}
%\end{equation}
being $M$ and $N$ are the height and width of the images $\mathcal{I}$ and $P$ the number of channels (corresponding to the labels, \textit{i.e.} orientations \textit{x,y,z}).

In this work, we test four methods to compute the reference map $\mathcal{I}^{\mathcal{R}}$, three of them from the literature and one proposed in this paper.
Orientation Map \cite{Lee2009}, $\mathcal{I}^{OM}$ (Fig. \ref{fig:normal_lines}(c)), and Geometric Context \cite{Hedau2009}, $\mathcal{I}^{GC}$ (Fig.~\ref{fig:normal_lines}(d)) are two methods widely used over years. Recently, researches \cite{PanoContext,jahromi2016geometric} have started to combine the strengths of both of them in one single map that we call Merge Map, $\mathcal{I}^{MM}$ (Fig. \ref{fig:normal_lines}(e)).

We propose a fourth method, \textit{Normal Map} ($\mathcal{I}^{NM}$), applying another recent deep learning method to our task. 
We choose the work from Eigen and Fergus \cite{Eigen2015}, which proposes a multiscale convolutional network that returns depth prediction, surface normal estimation and semantic labeling of indoor images. Here, we take advantage of the surface normal estimation to create the reference map. 
As expected, this network has been trained with conventional images, so it does not work properly with panoramic images.
To address this problem, we adapt the CNN to our image geometry by splitting the panorama in perspective images as in Section\ref{mall}. In this case, in order to stitch them back to the panorama, we need to rotate the normals to set them in a common reference frame. 
% The fourth method, \textit{Normal Map} ($\mathcal{I}^{NM}$) is a new reference map we introduce, that applies another recent deep learning method to our task and image geometry.
% In particular, we adapted the work from Eigen \textit{et al.} ~\cite{Eigen2015}, which proposes a multiscale convolutional network that returns depth prediction, surface normal estimation and semantic labeling of indoor images.
% We take advantage of the surface normal estimation to create the reference map.
% However, since this network has been trained from images coming from a RGB-D camera, it does not work properly with panoramic images, as expected.
% To address this problem, we split the panorama in perspective images as in Section\ref{mall}. In this case, in order to stitch perspective images back to the panorama we need to rotate the normals to set them in a common reference frame. 
%Two rotations are carried out: a first rotation associated with that initially performed to generate the perspective image from the original panorama (with the coordinates of the center points of each image) and one second rotation associated with the scene VPs. 
Overlapping areas are tackled in this case by doing the per-pixel average to achieve a continuity of the overall image.
Then we apply an angular threshold to determine whether or not the normals from each pixel belong to a main direction (VPs) and label them accordingly.
Resulting normal map is shown in Fig. \ref{fig:normal_lines}(b). It can be noticed that the ceiling is the worst estimated part by the CNN since black pixels means uncertain areas (\textit{i.e.} not belonging to any main direction). This happens because the CNN was trained with images where ceiling does not usually appear, making it difficult for the net to predict them.

\section{Experiments}
\begin{figure}%*} 
\begin{center}
     \subfloat{\includegraphics[width=1\linewidth]{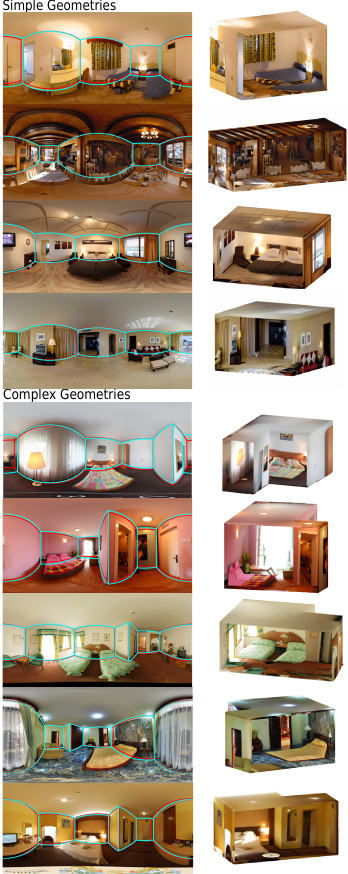}} 
\end{center}
\caption{Final layout estimations handling \textbf{different geometries} (cyan) compared with their ground truth (red).}
\label{fig:experi2}
\end{figure}

We have evaluated our proposal using full-view panoramas of indoor scenarios from two public datasets.
In particular, most of our quantitative results have been obtained from a subset of 85 panoramas of bedrooms and living rooms of the SUN360 dataset \cite{Xiao2012}.
Additionally, we also show some results using the Stanford (2D-3D-S) dataset \cite{2017arXiv170201105A}. 
For each panorama we have manually created the ground truth as a labeled image $\mathcal{I}^{GT}$, similar to those in Fig.~\ref{fig:normal_lines}, where each pixel encodes the direction of the surface it belongs to. 
A previous ground truth was provided by \cite{PanoContext}, but was unusable for us since images were labeled following the box-shaped rooms simplification.
The accuracy of our results is evaluated by computing $EOP\left(\mathcal{I}^{H_b},\mathcal{I}^{GT}\right)$, measuring the ratio of equally-oriented pixels between the best hypothesis and the ground truth.
Each EOP value shown is a median of 10 times performing the experiment. 
The number of hypotheses drawn ($N_h$) is specified in each experiment. For the experiments we allow the algorithm to initially select from three to five corners, \textit{i.e.} to solve layouts with four to eight walls.
Some examples of final layout estimations and 3D models are shown in Fig.~\ref{fig:experi2}.
This submission includes a video which illustrates the procedure and shows some additional results.
%We have evaluated our proposal in a subset $Np=$ 110 full-view panoramas in total of indoor scenarios from two public datasets, SUN360 \cite{Xiao2012} and Stanford (2D-3D-S) \cite{2017arXiv170201105A}. See final layout estimations and 3D models in Fig.\ref{fig:experi2}. This paper includes a video which shows the potential of our method including examples and illustrating the procedure. We use the Ground Truth, $\mathcal{I}^{GT} = \lbrace \mathcal{I}^{GT}_{1},...,\mathcal{I}^{GT}_{Np} \rbrace$, manually labeled by ourselves, in which each pixel in the image is labeled according to the direction of the surface it belongs to, considering as surfaces: ceiling, floor and walls. The accuracy of our results is evaluated by computing $EOP\left(\mathcal{I}^{H_b},\mathcal{I}^{GT}\right)$, measuring in this case the ratio of equally-oriented pixels between the labeled image given by the best final hypothesis and the ground truth.
%Each result shown is a median of 10 times performing the experiment and the hypotheses number drawn $N_h$ are specified in each experiment. 

% TABLA
\begin{table} %[h]
\centering
\begin{tabular}{ccc}
\toprule
 & \textit{EOP} & \textit{Computing Time} \\
 \midrule
Normal Map ($\mathcal{I}^{NM}$) & \textbf{0.925$\pm$0.061} &  243.36$\pm$1.42 \\
Orientation Map ($\mathcal{I}^{OM}$) &  0.906$\pm$0.133 & \textbf{23.54$\pm$4.16} \\
Geometric Context ($\mathcal{I}^{GC}$) & 0.883$\pm$0.114 & 174.07$\pm$13.28 \\
Merge Map ($\mathcal{I}^{MM}$) &0.923$\pm$0.147 & 197.61$\pm$17.44  \\
\bottomrule
\end{tabular}
\caption{\label{tab:referenceMaps}Ratio of equally-oriented pixels when comparing the best final hypotheses, $\mathcal{I}^{H_b}$, with the ground truth $\mathcal{I}^{GT}$, evaluating in each case with a \textit{reference map}. Also the computing time in seconds of generating each map is shown.}
\end{table}

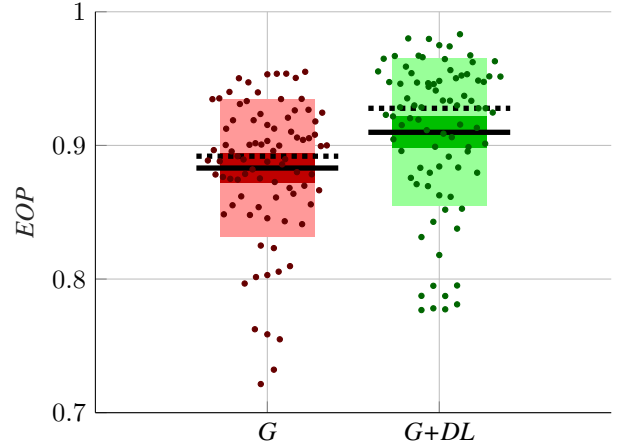
\begin{figure} 
\begin{tikzpicture}

\begin{axis}[%
width=2.7in,
height=2.1in,
at={(0.642in,0.544in)},
scale only axis,
xmin=0,
xmax=3,
xtick={1,2},
xticklabels={\textit{G}, \textit{G+DL}},
xlabel style={font=\color{white!15!black}},
ymin=0.7,
ymax=1,
ylabel style={font=\color{white!15!black}},
ylabel={\textit{EOP}},
axis background/.style={fill=white},
title style={font=\bfseries},
axis x line*=bottom,
axis y line*=left,
xmajorgrids,
ymajorgrids
]

\addplot[area legend, draw=none, fill=white!60!red, forget plot]
table[row sep=crcr] {%
x	y\\
0.725	0.831588666554601\\
1.275	0.831588666554601\\
1.275	0.934546048773932\\
0.725	0.934546048773932\\
}--cycle;

\addplot[area legend, draw=none, fill=black!25!red, forget plot]
table[row sep=crcr] {%
x	y\\
0.725	0.872123407554548\\
1.275	0.872123407554548\\
1.275	0.894011307773985\\
0.725	0.894011307773985\\
}--cycle;

\addplot [color=black, line width=2.0pt, forget plot]
  table[row sep=crcr]{%
0.5875	0.883067357664267\\
1.4125	0.883067357664267\\
};
\addplot [color=black, dotted, line width=2.0pt, forget plot]
  table[row sep=crcr]{%
0.5875	0.891934467548834\\
1.4125	0.891934467548834\\
};
\addplot [color=black!60!red, draw=none, mark size=1.0pt, mark=*, mark options={solid, fill=white!40!red, black!60!red}, forget plot]
  table[row sep=crcr]{%
0.680758607197862	0.934609021995171\\
0.720663781298129	0.935203990411071\\
0.654134113294645	0.888709226653891\\
0.779731585646889	0.940027096883945\\
0.760568955398397	0.912496680154353\\
0.747151776468577	0.848431904541202\\
0.834798689235166	0.950252221475799\\
0.688720701965181	0.896494590170479\\
0.723307290635716	0.888116428071747\\
0.800474129498664	0.918820443652889\\
0.698638785576642	0.878239344814358\\
0.757893879306252	0.900310459813811\\
0.741690387637122	0.876392816287987\\
0.792480467976787	0.895494730775707\\
0.784741989697602	0.87500195285038\\
0.827793591758081	0.874399173032959\\
0.827067056647322	0.891934467548834\\
0.868128018414256	0.796654463521623\\
0.861653645317858	0.889473008136009\\
0.870845193818561	0.87866202843003\\
0.896240233988393	0.900467121810995\\
0.934064009207128	0.801497098498301\\
0.913896795879041	0.882042195455153\\
0.930826822658929	0.901635794271986\\
0.797721421174862	0.855219144530023\\
0.965413411329465	0.900650689746754\\
0.961934967214384	0.721422421499754\\
1	0.903329132535182\\
0.840379303598931	0.930990607223637\\
1.03458658867054	0.895897451920824\\
0.880284477699198	0.933176063782697\\
0.848291065881146	0.861872288792722\\
1.06917317734107	0.899854360758296\\
0.920189651799466	0.918861670494253\\
0.898860710587431	0.847935012611074\\
0.949430355293715	0.853799639286836\\
0.961934967214384	0.825061927055397\\
0.95694839793952	0.875310937177237\\
1	0.88782176464768\\
1	0.845490911868296\\
1.10375976601161	0.889691727378615\\
1.03806503278562	0.732188268663608\\
0.927492301231193	0.76244139279025\\
1.05056964470628	0.861048619898938\\
1	0.758628126947426\\
1.03806503278562	0.823159850784872\\
1.13834635468214	0.910234845447081\\
1.17293294335268	0.905888234467028\\
1.07250769876881	0.754912069656871\\
1.04305160206048	0.872624682987288\\
0.889865792823444	0.947183208611289\\
0.944932896411722	0.939672546048212\\
1	0.953175855478657\\
0.960094825899733	0.923529850836948\\
1	0.915324841471946\\
1.03990517410027	0.912573058302565\\
1.05506710358828	0.953648011303966\\
1.11013420717656	0.953697049546852\\
1.07981034820053	0.934684098243129\\
1.20751953202321	0.90413631069242\\
1.10113928941257	0.84325294533234\\
1.16520131076483	0.950497846656981\\
1.1197155223008	0.920756803296759\\
1.15962069640107	0.926225218328673\\
1.22026841435311	0.955105271654507\\
1.19952587050134	0.934820363803007\\
1.15170893411885	0.86384423371019\\
1.08610320412096	0.885808592888847\\
1.2394310446016	0.926670902182158\\
1.12915480618144	0.868159599084157\\
1.17220640824192	0.880024250062057\\
1.27933621870187	0.917936019413937\\
1.2152580103024	0.869819087940758\\
1	0.802965641984374\\
1.06593599079287	0.80552474391622\\
1.24210612069375	0.905389606669895\\
1.31924139280214	0.924503672226649\\
1.27669270936428	0.908025954683456\\
1.31127929803482	0.899463790682213\\
1.25830961236288	0.878376478307738\\
1.30136121442336	0.866435449181626\\
1.20227857882514	0.841067054806529\\
1.34586588670536	0.899998871686447\\
1.25284822353142	0.855883547626115\\
1.13187198158574	0.809666522589705\\
};

\addplot[area legend, draw=none, fill=white!60!green, forget plot]
table[row sep=crcr] {%
x	y\\
1.725	0.854702419663112\\
2.275	0.854702419663112\\
2.275	0.964933125869345\\
1.725	0.964933125869345\\
}--cycle;

\addplot[area legend, draw=none, fill=black!25!green, forget plot]
table[row sep=crcr] {%
x	y\\
1.725	0.898100697960362\\
2.275	0.898100697960362\\
2.275	0.921534847572094\\
1.725	0.921534847572094\\
}--cycle;

\addplot [color=black, line width=2.0pt, forget plot]
  table[row sep=crcr]{%
1.5875	0.909817772766228\\
2.4125	0.909817772766228\\
};
\addplot [color=black, dotted, line width=2.0pt, forget plot]
  table[row sep=crcr]{%
1.5875	0.927781423098488\\
2.4125	0.927781423098488\\
};
\addplot [color=black!60!green, draw=none, mark size=1.0pt, mark=*, mark options={solid, fill=white!40!green, black!60!green}, forget plot]
  table[row sep=crcr]{%
1.81952465443761	0.980053152247687\\
1.68925206405937	0.92287412707588\\
2	0.817969174473729\\
1.64432126334493	0.955365651705229\\
1.67665569394994	0.964866485789325\\
1.77973158564689	0.89583496070865\\
1.70899012455495	0.947394550419125\\
1.87968310295841	0.967155660401923\\
1.86812801841426	0.870997307670276\\
1.73314843347923	0.90460152304971\\
1.83479868923517	0.875717997989866\\
1.78166690011937	0.915217651684399\\
1.88986579282344	0.883246887156494\\
1.73068512218479	0.921711530149406\\
1.74132455515995	0.966904827619727\\
1.9398415514792	0.979683412575662\\
1.77365898576496	0.946144292208908\\
1.89632728827269	0.787427462457536\\
1.94493289641172	0.879558603738016\\
1.77211818031021	0.928500939971983\\
1.81355123843562	0.935220047180865\\
1.96544242942423	0.794953313856906\\
1.83018536675951	0.920337591415096\\
1.87870383339965	0.919320807317027\\
1.92722230003979	0.911420442611369\\
2	0.884233293581979\\
1.80599341636996	0.958957160538188\\
1.97574076667993	0.898982521555129\\
1.83832784697497	0.954058109883853\\
1.85498429656104	0.929917407447911\\
1.87066227757998	0.947027414547607\\
1.89641735468646	0.928505713606246\\
2.05506710358828	0.891591633815384\\
1.89632728827269	0.831433426896565\\
2.02425923332007	0.908781490797301\\
1.96544242942423	0.842835469317683\\
2.03455757057577	0.851915789619862\\
2.07277769996021	0.906147312617496\\
1.93785041281187	0.943906325672952\\
1.90299670818498	0.966003478677478\\
2.12129616660035	0.915619504896013\\
1.89632728827269	0.776794409119551\\
1.96544242942423	0.778088931938391\\
1.93533113878999	0.946563504090571\\
2.03455757057577	0.777349018627589\\
2.10367271172731	0.837737227924545\\
1.96766556939499	0.945887817858947\\
2	0.948256408387015\\
2.11013420717656	0.88347645556788\\
2.16520131076483	0.89549603267596\\
1.97928347093729	0.94108901352414\\
2.03233443060501	0.96459786037033\\
2	0.974948401353282\\
2.02071652906271	0.933587030296087\\
2.06214958718813	0.933463349771994\\
2.06466886121001	0.956342076895436\\
1.93406400920713	0.869523556583189\\
2.0601584485208	0.97409305288666\\
2.09700329181502	0.950266542378589\\
2.12933772242002	0.952307488009499\\
2.16167215302503	0.953285649066711\\
2.12031689704159	0.983184656324371\\
2.19400658363004	0.962303912123469\\
2.10358264531354	0.929944747353237\\
2.14501570343896	0.937119085684131\\
2.22634101423504	0.9485393547088\\
2.25867544484005	0.947524740444486\\
2.29100987544505	0.951667387051474\\
2.18644876156438	0.933367877086729\\
2.22788181968979	0.927788800533258\\
2.18047534556239	0.967394342115085\\
2.26931487781521	0.927781423098488\\
2.22026841435311	0.879575962408064\\
2	0.862657334645649\\
2.03455757057577	0.787322876470496\\
2.31074793594063	0.924600880778918\\
2.32334430605006	0.962974390754078\\
2.35567873665507	0.951512026954543\\
2.16981463324049	0.898788104450589\\
2.10367271172731	0.781052056915607\\
2.21833309988063	0.913174970186484\\
2.06593599079287	0.861538134394295\\
2.26685156652077	0.901319432510359\\
2.10367271172731	0.795182448301541\\
2.13187198158574	0.852647457562391\\
};
\end{axis}
\end{tikzpicture}% 
\caption{\label{boxplot}\textbf{Advantages of combining.} Here we highlight the advantages of using structural lines from Geometry and Deep Learning combination \cite{Mallya:2015} over lines obtained only with Geometry. The mean is represented in solid black and the median in dotted black. Also the standard deviation is shown in light color and jittered raw data are plotted for each group.}
\end{figure}

\paragraph{\textbf{Edge map advantages}} A comparative study showing the effects of selecting structural lines (Section \ref{mall}) can be found in Fig. \ref{boxplot}. 
For this experiment we choose $N_h=$ 100 and the $\mathcal{I}^{NM}$ as reference map. 
Every single image evaluated is represented as a point: green if geometry and deep learning (G+DL) are used to obtain structural lines, and red if just geometry (G) is used.
The graph demonstrates the improvement when combining both techniques, highlighted by the mean and especially median values: 0.889 vs. 0.925.
%Mean and especially median values of EOP highlight the improvement: 0.889 vs. 0.925. 
Thus, the experiment proves that the inclusion of DL techniques in the pipeline of the process clearly benefits the approach.
In particular, the detection of structural lines allows to remove clutter effectively, which translates into better accuracy.
%With this we want to draw attention to the benefits of our approach, which exploits geometry and deep learning combination to select structural lines leading to better layout hypotheses.

\paragraph{\textbf{Reference maps comparison}}
We compare the performance using the four alternative reference maps at hypotheses evaluation step (Section~\ref{hev}). %$\mathcal{I}^{R}$, which provide pixel-wise surface normal estimation. 
Here, we use $N_h=$ 100 as well. 
%Fig.\ref{fig:normal_lines}(b)-(e) contains a visual example of how these reference maps look like. 
Table \ref{tab:referenceMaps} shows the median EOP value and the computing time of creating each map.
In terms of accuracy, $\mathcal{I}^{NM}$ and $\mathcal{I}^{MM}$ perform similarly in median, although the smaller standard deviation of the $\mathcal{I}^{NM}$ indicates more consistent results.
Both are considerably better than $\mathcal{I}^{OM}$ and $\mathcal{I}^{GC}$.
However, the $\mathcal{I}^{OM}$ is about ten times faster to compute than the $\mathcal{I}^{NM}$ and, therefore, its usage would be recommendable if the priority lies in getting fast results in spite of losing some accuracy.
The smaller standard deviation on the computing time of the $\mathcal{I}^{NM}$ shows that it does not vary through images, unlike the others whose time depends on scene-specific features such as the number of lines.
%For a first application of deep learning in these task, the results are quite favorable
%but, if we pay special attention to the standard deviation of the data, $\mathcal{I}^{NM}$ outperforms $\mathcal{I}^{MM}$. We also collected data of computing times and show that generate the $\mathcal{I}^{NM}$ is slower so, depending on the priority between accuracy or time, we would recommend to use $\mathcal{I}^{NM}$ or $\mathcal{I}^{OM}$ respectively. Also with this metric we appreciate that the $\mathcal{I}^{NM}$ has the lowest standard deviation whereas the $\mathcal{I}^{MM}$ has the highest. That means that the difference between their computing times is not very significant.

\begin{figure} 

\begin{tikzpicture}

\begin{axis}[%
width=2.5in,
height=2in,
at={(0.772in,0.516in)},
scale only axis,
xmin=10,
xmax=100,
xlabel style={font=\color{white!15!black}},
xlabel={Number of hypotheses evaluated},
axis y line*=left,
ymin=0.80,
ymax=0.95,
ylabel style={font=\color{black}},
ylabel={\textit{EOP}},
%axis background/.style={fill=white},
yticklabel pos=right,
legend style={at={(0.43,0.23)}, anchor=south west, legend cell align=left, align=left, draw=white!15!black}
]
\addplot+[line width=1.5pt,mark size=2pt][color=cyan, mark=otimes*, mark options={solid, cyan}]
  table[row sep=crcr]{%
5	0.899810530116425\\
10	0.91256047326678\\
20	0.921697209246616\\
40	0.929944747353237\\
60	0.929953860655012\\
80	0.933160874946405\\
100	0.933471161173516\\
}; \label{Our_EOP}
\addlegendentry{\small Our EOP}
\addplot+[line width=1.5pt,mark size=2pt][color=blue, mark=otimes*, mark options={solid, blue}]
  table[row sep=crcr]{%
5	0.826551734912278\\
10	0.829366009293832\\
20	0.853505843796272\\
40	0.884003074220466\\
60	0.890187751375241\\
80	0.895196161650879\\
100	0.896562072000292\\
};\label{PanoContext_EOP}
\addlegendentry{\small PanoContext EOP}
\end{axis}

\begin{axis}[%
width=2.5in,
height=2in,
at={(0.772in,0.516in)},
scale only axis,
xmin=10,
xmax=100,
xlabel style={font=\color{white!15!black}},
xlabel={Number of hypotheses evaluated},
axis y line*=right,
ymin=0,
ymax=10,
ylabel style={font=\color{black}},
ylabel={Computing Time (s)},
%axis background/.style={fill=white},
yticklabel pos=right,
legend style={at={(0.43,0.03)}, anchor=south west, legend cell align=left, align=left, draw=white!15!black, fill=lightgray}
]
%\addlegendimage{/pgfplots/refstyle=Our_EOP}\addlegendentry{\small Our EOP}
%\addlegendimage{/pgfplots/refstyle=PanoContext_EOP}\addlegendentry{\small PanoContext EOP}
\addplot+[line width=0.8pt][color=cyan, dashed, mark=asterisk, mark options={solid, cyan}]
  table[row sep=crcr]{%
5	0.96949\\
10	1.490578\\
20	2.536239\\
40	4.62271\\
60	6.591172\\
80	7.82\\
100	9.31\\
};
\addlegendentry{\small Our CT}
\addplot+[line width=0.8pt][color=blue, dashed, mark=asterisk, mark options={solid, blue}]
  table[row sep=crcr]{%
5	6.06394508333333\\
10	6.10611158333333\\
20	6.52383133333333\\
40	6.75799458333333\\
60	7.02790558333333\\
80	7.2423385\\
100	7.48749558333333\\
};
\addlegendentry{\small PanoContext CT}
\end{axis}

\end{tikzpicture}% 
\caption{\label{N_PC} \textbf{Comparison with PanoContext} \cite{PanoContext} (with only four-wall rooms). We show the ratio of equally-oriented pixels and computing time against the number of hypotheses. Our method outperforms PanoContext and is able to provide much better results and much faster with fewer hypotheses.}
\end{figure}
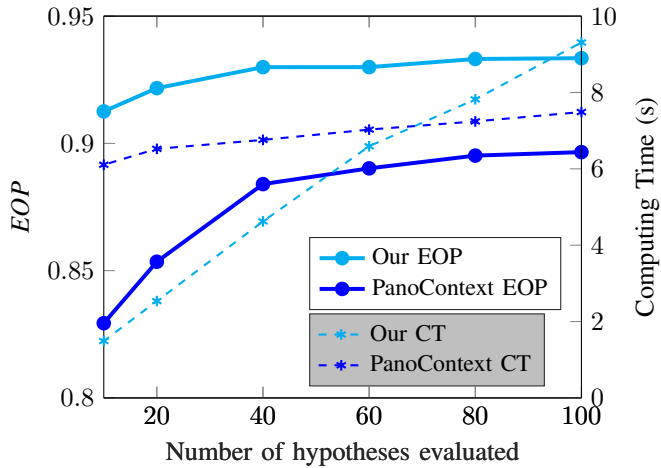

\begin{figure} 
\begin{center}
%    \subfloat{\includegraphics[width=1\linewidth]{ComplexGeo2.pdf}} 
   \subfloat{\includegraphics[width=0.5\linewidth]{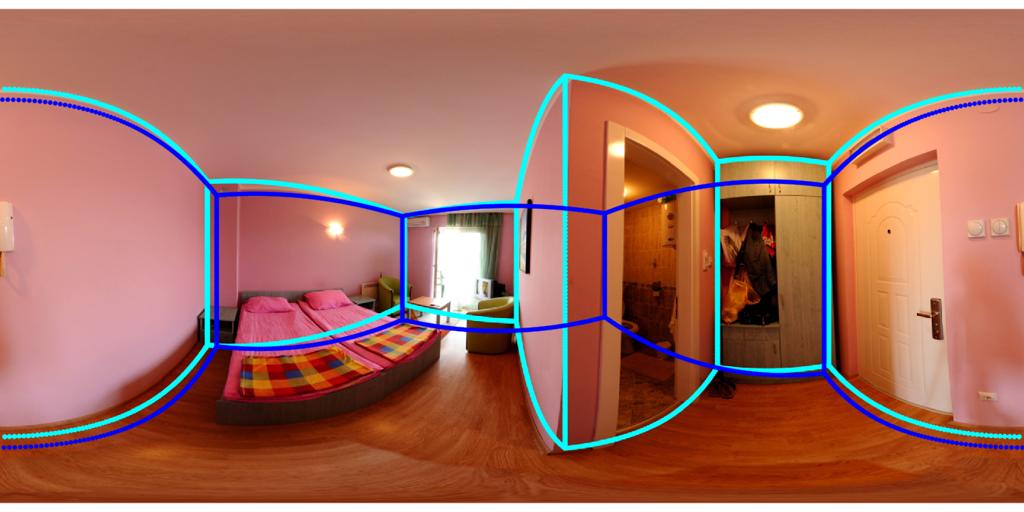}} \hspace*{0.003\linewidth}
    \subfloat{\includegraphics[width=0.5\linewidth]{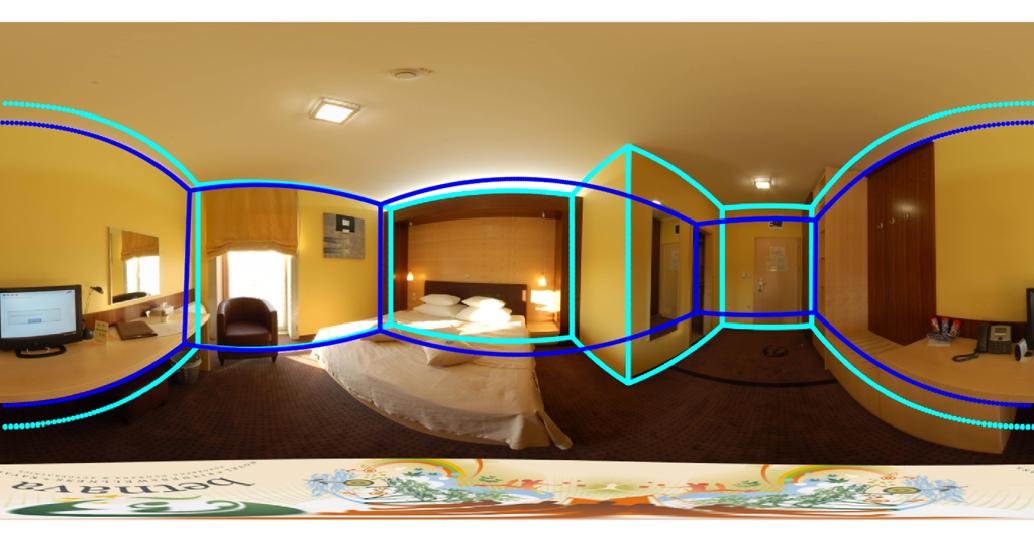}}
\end{center}
   \caption{\textbf{Comparison with PanoContext} \cite{PanoContext} in complex geometries. Our method (cyan) is able to find 6 walls whereas \cite{PanoContext} (dark blue) always finds just 4 walls.}
   \label{fig:PCyOurs}
\end{figure}

 \paragraph{\textbf{Comparison with the state of the art}}
We perform a comparison with PanoContext \cite{PanoContext} since it is, to our knowledge, the only directly related method with available code.
%At the end, our algorithm in comparison to another very similar state of the art method is evaluated and shown in Fig.\ref{N_PC}, the \textit{PanoContext} ~\cite{PanoContext}, $PC$. 
%The method starts with the estimation of the layout of the room and then performs object detection as well. Since our method only deals with the first task we use their initial results
We establish the comparison with the first stage of their algorithm that reaches the same point as our work does, since after layout extraction they introduce object detection in the method.
To evaluate accuracy, in order to carry out a direct and fair comparison we only compare numerically the four-wall rooms cases, removing more complex shaped ones from the experiment.
%As we have mentioned before, we manually labeled our $\mathcal{I}^{GT}$. \textit{PanoContext} made also a Ground Truth with the same dataset but it was not useful for our accuracy evaluation since all cases were considered as four-wall rooms whereas our approach is not constrained to that case. However, in order to carry out a direct and fair comparison with the state of the art, we only compare numerically our four-wall rooms cases with theirs, without considering our more complex shaped ones.
In Fig.~\ref{N_PC} we show the EOP ratio and the computing time necessary to generate the hypotheses for each method, varying the number of hypotheses $N_h$.
Our method clearly outperforms \cite{PanoContext}, being the difference larger when only a few hypotheses are considered. Although the difference decreases as the amount of hypotheses rises, when both methods reach a stable EOP value our proposal continues giving better results. Moreover, with just 10 hypotheses (91,26\%) our method beats \cite{PanoContext} with 100 hypotheses (89.66\%). This shows the good performance of our structural lines selection which increases the likelihood of getting good hypotheses with only a few attempts. Computing times show again bigger difference when fewer hypothesis are evaluated. Only rooms up to 4 walls are considered here to be fair with \cite{PanoContext}, but our method is also able to deal with more complex rooms (see  Fig.~\ref{fig:PCyOurs}).
%Besides, some scenes have more than four walls, which our method is able to solve unlike theirs. %Fig.~\ref{fig:PCyOurs} shows an example of the results given by both methods for more complex shaped rooms.
% \begin{figure} 
% \input{N_PC_si.tex} 
% \caption{\label{N_PC} Comparison with PanoContext \cite{PanoContext} (with only four walled rooms). We show the Pixel Accuracy against the number of hypotheses. Our method outperforms PanoContext and is able to provide much better results with fewer hypotheses.}
% \end{figure}

% \begin{table} [t]
% \begin{center}
% \begin{tabular}{c c c c c c c}%{\linewidth}{l*{6}{c}}
% \toprule
% Method & Dataset & \multicolumn{5}{c}{Number of hypotheses evaluated $Nh$} \\ %20it & 40it & 60it & 80it & 100it \\ 
% \addlinespace
% & & 20 & 40 & 60 & 80 & 100 \\ 
% \midrule
% Ours & LSUN360 & \textbf{2.36} & \textbf{4.31} & \textbf{5.91} & 7.82 & 9.31 \\
% %\midrule
% PC \cite{PanoContext} & LSUN360 & 6.52 & 6.76 & 7.03 & \textbf{7.24}& \textbf{7.49} \\
% \bottomrule
% \end{tabular}
% \end{center}
% \caption{\label{tab:Time} Computing time in seconds at layout hypotheses evaluation step}
% \end{table}

\begin{figure}
\begin{center}
	\subfloat{\includegraphics[width=0.97\linewidth]{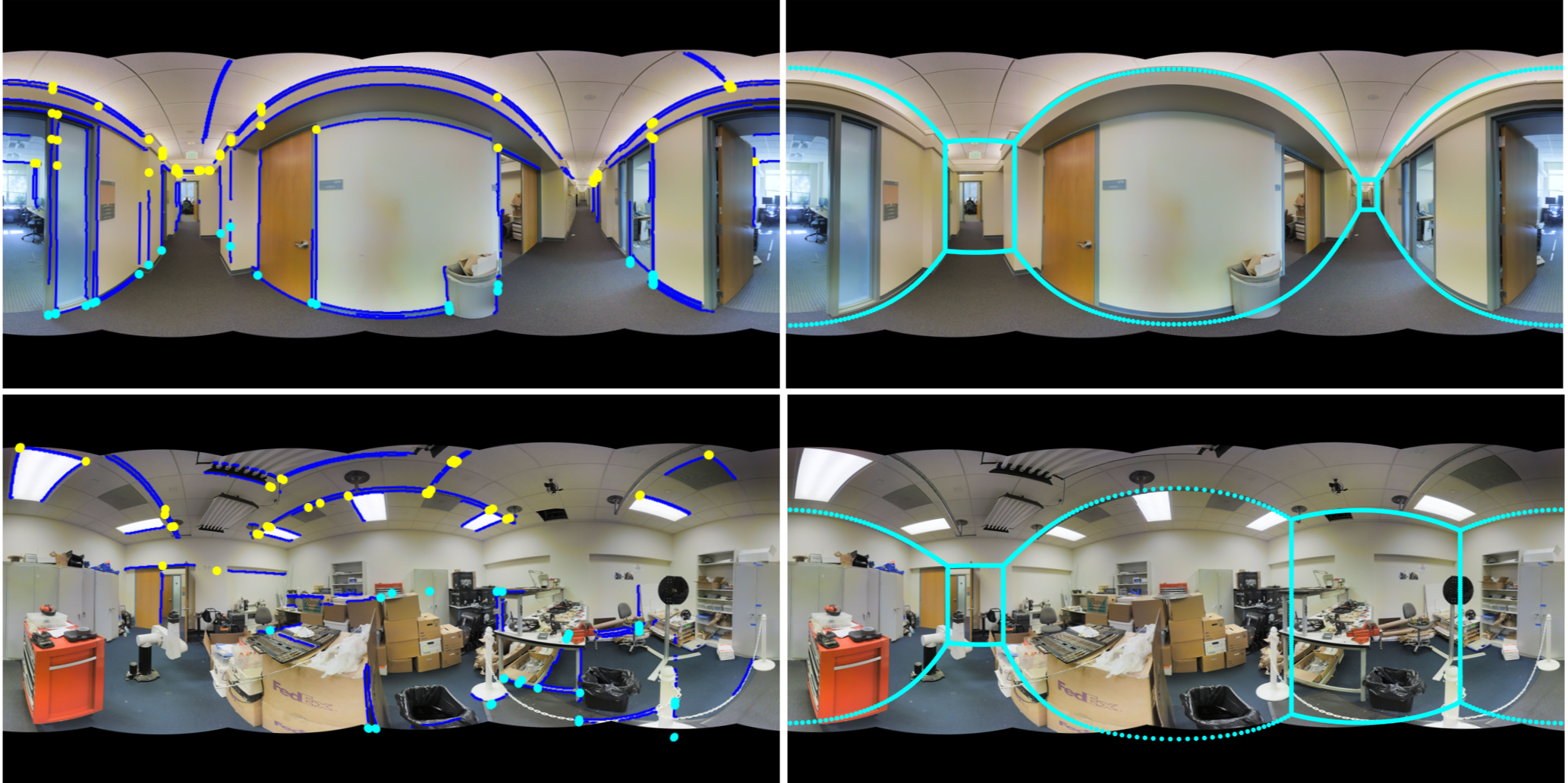}}
%    \subfloat{\includegraphics[width=0.5\linewidth]{21_corners.png}{\label{fig:corners1}}} \hspace*{0.003\linewidth}
%     \subfloat{\includegraphics[width=0.5\linewidth]{21_layout.png}{\label{fig:corners2}}}\\
%      \subfloat{\includegraphics[width=0.5\linewidth]{19_corners.png}{\label{fig:corners1}}} \hspace*{0.003\linewidth}
%     \subfloat{\includegraphics[width=0.5\linewidth]{19_layout.png}{\label{fig:corners2}}}
\end{center}
   \caption{\textbf{Top}: challenging corridor well estimated by our approach in Stanford (2D-3D-S) dataset. \textbf{Bottom}: a clear case of failure.}
   \label{fig:standf}
\end{figure}

\paragraph{\textbf{Different datasets}}
%The SUN360 dataset is composed of $\thicksim$700 panoramas covering 360$^\circ$ horizontal and 180$^\circ$ vertical FOV including bedrooms and living-rooms from which we evaluate 85, whereas 
Besides the 85 images from the SUN360 dataset, we additionally tested our method with 25 panoramas from the Stanford (2D-3D-S) dataset. %, which includes panoramas of 6 indoor areas from 3 different buildings of educational and office use. The criteria to select panoramas has been choosing scenes that satisfy the Manhattan world assumption. 
In Table \ref{tab:StvsLsun} we show the EOP we reach in both datasets. Several reasons can be associated with the fact that our proposal works better with SUN360 dataset. On the one hand, panoramas from the Stanford dataset do not cover full view vertically, leaving a black mask that can lead to confusions in the limits when extracting structural lines. On the other hand, indoor scenes represented in the second dataset show more challenging scenarios like cluttered laboratories or corridors instead of bedrooms and living-rooms (see Fig.~\ref{fig:standf}). Still, our method achieves more than 87$\%$ of Equally Oriented Pixels in this dataset.
\begin{table}
\begin{center}
\begin{tabular}{c c c }%{\linewidth}{l*{6}{c}}
\toprule
Dataset & Category & EOP ($N_h=$100) \\ %20it & 40it & 60it & 80it & 100it \\ 
\midrule
\multirow{ 2}{*}{LSUN360} &  bedroom & 0.921\\ & livingroom & 0.933  \\
%\midrule
\multirow{ 2}{*}{Stanford (2D-3D-S)} & area1 & 0.873  \\ & area3 & 0.885 \\
\bottomrule
\end{tabular}
\end{center}
\caption{\label{tab:StvsLsun} Ratio of equally-oriented pixels evaluated in different scenarios from two public datasets.}
\end{table}

\section{Conclusion}
We propose a novel entire pipeline which converts 360$^\circ$ panoramas
into flexible, closed, 3D reconstructions of the rooms represented in the images.
%In this work we exploit the 360$^{\circ}$ full-view of single panoramas to estimate 3D room layouts under Manhattan World assumption. We propose a method that fuses the accuracy of geometric reasoning based on lines and the abstraction reached with deep learning techniques. We obtain structural corners from which, with very few hypotheses, we achieve flexible closed geometries without shape assumptions. 
Our experimental results show that the proposed algorithm has a good performance in scene interpretation of full-view images and outperforms the state of the art not only in terms of accuracy but also in speed. As future work we consider to train a CNN able to work with both conventional and omnidirectional images.
% In this work we introduce the main novelty of the exploitation of Deep Learning approaches for single panoramic images applied to the problem of 3D room layout estimation with Manhattan World assumption (without four-walls simplifications), for which we propose a new flexible method that integrates old and new techniques fusing geometric reasoning in computer vision with two different deep neural networks \cite{Mallya:2015,Eigen2015} adapted to the proposed image geometry.  
% Our experimental results imply that the proposed algorithm has a good performance in scene interpretation of full-view images and overcomes the state of the art.

%*}

{\small
\bibliographystyle{ieee}
\bibliography{egbib}
}

\end{document}